\documentclass[lettersize,journal]{IEEEtran}
\usepackage{amsmath,amsfonts}
\usepackage{algorithm}
\usepackage{array}
\usepackage{textcomp}
\usepackage{stfloats}
\usepackage{url}
\usepackage{verbatim}
\usepackage{graphicx}
\usepackage{cite}
\usepackage{mathrsfs}
\usepackage{tabularray}
\usepackage{xcolor}
\definecolor{lightgray}{RGB}{217, 217, 217}
\definecolor{forestgreen}{RGB}{47, 159, 87}%
\definecolor{forestred}{RGB}{202,12,22}%
\definecolor{lightpink}{HTML}{F9E6EC}
\usepackage{wrapfig}
\usepackage{booktabs} 
\usepackage{algpseudocode}
\usepackage{subcaption}
\definecolor{SGDColor}{RGB}{0,82,147}  % 定义SGD标准蓝
\definecolor{CFlatColor}{RGB}{213,0,50} % 定义C-Flat标准红
\usepackage{newtxtext, newtxmath}

\usepackage{tikz,hyperref}
\DeclareRobustCommand{\orcidicon}{
\begin{tikzpicture}
\draw[lime, fill=lime] (0,0)
circle[radius=0.16]
node[white]{{\fontfamily{qag}\selectfont \tiny \.{I}D}}; 
\end{tikzpicture}
\hspace{-2mm}
}
\foreach \x in {A, ..., Z}{%
\expandafter\xdef\csname orcid\x\endcsname{\noexpand\href{https://orcid.org/\csname orcidauthor\x\endcsname}{\noexpand\orcidicon}}
}
\begin{document}

\title{C-Flat++: Towards a More Efficient and Powerful Framework for Continual Learning}

\author{Wei Li\hspace{-.5mm}\orcidA{}, Hangjie Yuan, Zixiang Zhao, Yifan Zhu, Aojun Lu, Tao Feng*\hspace{-.5mm}\orcidB{}, Yanan Sun
\thanks{*Corresponding authors: Tao Feng \\
Wei Li, Aojun Lu and Yanan Sun are with the College of Computer Science, Sichuan University, China. (E-mail: ymjiii98@gmail.com; aojunlu@stu.scu.edu.cn; ysun@scu.edu.cn) \\
Hangjie Yuan is with Zhejiang University, Zhejiang, China. (E-mail: hj.yuan@zju.edu.cn) \\
Zixiang Zhao is with ETH Zürich, Switzerland. (E-mail: zixiang.zhao@ethz.ch) \\
Yifan Zhu is with Beijing University of Posts and Telecommunications, China. (E-mail: yifan\_zhu@bupt.edu.cn) \\
Tao Feng is with the Department of Computer Science and Technology, Tsinghua University, China. (E-mail: fengtao.hi@gmail.com)
}}



\maketitle

\begin{abstract}
% How to balance the learning ’sensitivity-stability’ upon new task training and memory preserving is critical in CL to resolve catastrophic forgetting. 
% Improving model generalization ability within each learning phase is one solution to help CL learning overcome the gap in the joint knowledge space.
% Zeroth-order loss landscape sharpness-aware minimization is a strong training regime improving model generalization in transfer learning compared with optimizer like SGD.
% It has also been introduced into CL to improve memory representation or learning efficiency.
% However, zeroth-order sharpness alone could favors sharper over flatter minima in certain scenarios, leading to a rather sensitive minima rather than a global optima. 
% To further enhance learning stability, we propose a \textbf{C}ontinual \textbf{Flat}ness (\textbf{C-Flat}) method featuring a flatter loss landscape tailored for CL. C-Flat could be easily called with only one line of code and is plug-and-play to any CL methods. A general framework of C-Flat applied to all CL categories and a thorough comparison with loss minima optimizer and flat minima based CL approaches is presented in this paper, showing that our method can boost CL performance in almost all cases. Code is available at \url{https://github.com/WanNaa/C-Flat}.

Balancing sensitivity to new tasks and stability for retaining past knowledge is crucial in continual learning (CL). Recently, sharpness-aware minimization has proven effective in transfer learning and has also been adopted in continual learning (CL) to improve memory retention and learning efficiency. However, relying on zeroth-order sharpness alone may favor sharper minima over flatter ones in certain settings, leading to less robust and potentially suboptimal solutions.
In this paper, we propose \textbf{C}ontinual \textbf{Flat}ness (\textbf{C-Flat}), a method that promotes flatter loss landscapes tailored for CL. C-Flat offers plug-and-play compatibility, enabling easy integration with minimal modifications to the code pipeline. Besides, we present a general framework that integrates C-Flat into all major CL paradigms and conduct comprehensive comparisons with loss-minima optimizers and flat-minima-based CL methods. Our results show that C-Flat consistently improves performance across a wide range of settings.
In addition, we introduce C-Flat++, an efficient yet effective framework that leverages selective flatness-driven promotion, significantly reducing the update cost required by C-Flat. Extensive experiments across multiple CL methods, datasets, and scenarios demonstrate the effectiveness and efficiency of our proposed approaches.
Code is available at \url{https://github.com/WanNaa/C-Flat}.
\end{abstract}


\section{Introduction}
\label{sec:intro}

\IEEEPARstart{W}{hy} study Continual Learning (CL)? CL is generally acknowledged as a necessary attribute for Artificial General Intelligence (AGI)~\cite{hadsell2020embracing, feng2022identifying, masana2022class, zhou2023class}. In the open world, CL holds the potential for substantial benefits across many applications: \textit{e.g.} vision model needs to learn a growing image set~\cite{feng2022overcoming, Yuan2022RLIP, yuan2023rlipv2}, or, embodied model needs to incrementally add skills to their repertoire~\cite{driess2023palm, sun2025stronger, lu2025rethinking, zhang2025parameter}. 
% More fundamentally, CL can promote AGI by enhancing learning efficiency and facilitating knowledge transfer across related tasks.

\textbf{Challenges.} %The major challenge for CL lies in  
%%And this challenge presents an even tougher threat to class incremental learning (CIL), given the intricate nature of distributional changes involved.
A good CL model is expected to keep the memory of all seen tasks upon learning new knowledge~\cite{hadsell2020embracing}. 
However, due to the limited access to previous data, the learning phase is naturally sensitive to the current task, hence resulting in a major challenge in CL called catastrophic forgetting~\cite{delange2021continual}, which refers to the drastic performance drop on past knowledge after learning new knowledge.
This learning sensitivity-stability dilemma is critical in CL, requiring model with strong generalization ability~\cite{feng2022identifying} to overcome the
 knowledge gaps between sequentially arriving tasks.

\textbf{Current solutions.} A series of works~\cite{rebuffi2017icarl, NEURIPS2019_fa7cdfad, lin2023pcr, jeeveswaran2023birt} are proposed to improve learning stability by extending data space with dedicated selected and stored exemplars from old tasks, or frozen some network blocks or layers that are strongly related to previous knowledge~\cite{zhou2022model,hu2023dense,zhu2022self,DBLP:conf/cvpr/YanX021, hu2023dense}.

Another group of works seeks to preserve model generalization with regularisation onto the training procedure itself~\cite{DBLP:journals/pami/LiH18a, feng2022progressive, konishi2023parameter}.
Diverse weight~\cite{rudner2022continual, kim2022warping, akyurek2021subspace} or gradient alignment~\cite{hadsell2020embracing, delange2021continual, liu2022continual, jin2021gradient, liu2025branch} strategies are designed to encourage the training to efficiently extracting features for the current data space without forgetting.
%%A series of works are proposed to overcome forgetting, encompassing regularization, dedicated memory systems, and modular architectures. Another line that focuses on advancing CL from the source at the problem, namely, the gradient. For example, aligning gradients and mitigating conflicting gradients to overcome forgetting; Or, characterizing generalization from loss landscapes to promote CL. However, the study on the loss landscape is still in its infancy.

%%Taking a close look at these methods, we observe that they are the prelude to the investigation of flat minima in CL by demonstrating that sharpness-aware minimization could enhance CL. However, these efforts primarily take the zero-order sharpness into account, leading to a less-than-stellar performance. Second, their applicability is limited due to solely concern the loss landscape within the specific case, for example, FS-DGPM focuses on flatting sharpness in gradient memory projection. 

\begin{figure}
    \centering
    \includegraphics[width=.8\linewidth]{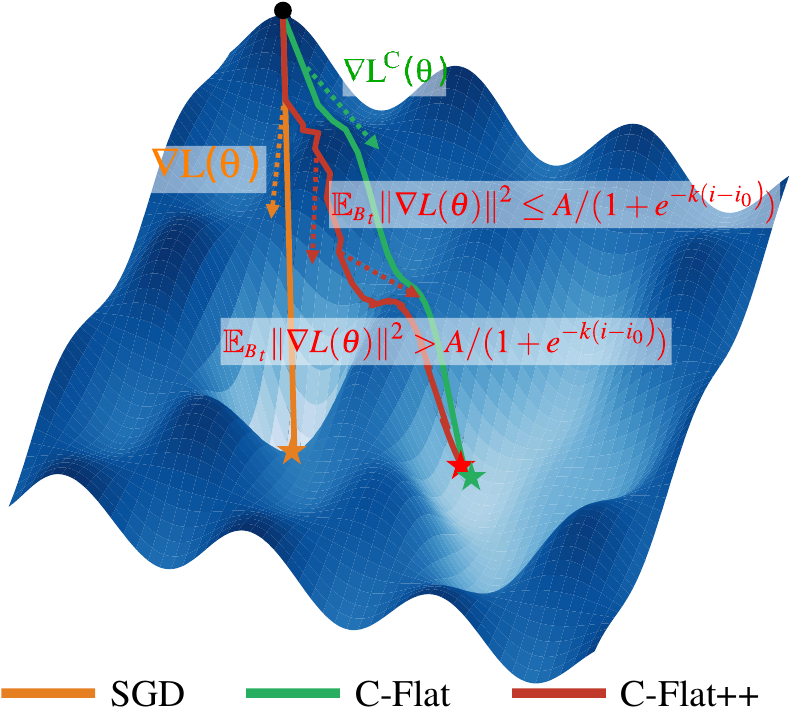}
    \caption{Considering a loss landscape with several sharp local minima, $\bullet$ and $\star$ denote the start and end points, respectively. \textcolor{orange}{SGD} follows the gradient direction \( \nabla L(\theta) \), often converging to sharp minima. \textcolor{forestgreen}{C-Flat} not only reduces the loss but also encourages exploration of flatter regions throughout optimization. \textcolor{red}{C-Flat++} selectively initiates flat-region exploration when the expected gradient norm \( \mathbb{E}_{B_t}\|\nabla L(\theta)\|^2 \) surpasses a sharpness-aware proxy of the loss surface geometry.}
    \label{fig:traj}
    \vspace{-4mm}
\end{figure}

\textbf{Loss landscape sharpness} optimization~\cite{he2019asymmetric, foret2020sharpness, zhong2022improving, zhuang2022surrogate} has recently gained attention~\cite{keskar2016large} as an efficient training strategy for improving model generalization. Conventional loss minimization optimizers such as SGD often lead to suboptimal solutions~\cite{baldassi2020shaping, liu2022towards, du2022sharpness} (see Figure~\ref{fig:traj}). 
To address this issue, zeroth-order sharpness-aware minimization, which seeks neighborhood-flat minima~\cite{DBLP:conf/iclr/ForetKMN21}, has shown strong potential in enhancing generalization, particularly in transfer learning tasks. This approach has also been adopted in several CL works~\cite{shi2021overcoming, kong2023overcoming} with task-specific designs to improve either old knowledge retention or few-shot learning efficiency.
However, given limited application scenarios~\cite{deng2021flattening, shi2021overcoming}, the zeroth-order sharpness optimization has been found to sometimes prefer sharper minima over truly flat solutions~\cite{zhuang2022surrogate}, which may still result in rapid gradient descent towards suboptimal regions.

\begin{figure*}
\centering
    \subfloat[Direct Tuning]{\includegraphics[width=1.2in]{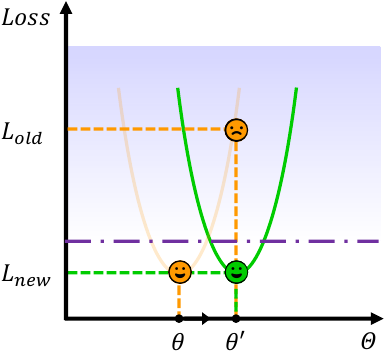}}
    \hfil
    \subfloat[Regularization]{\includegraphics[width=1.2in]{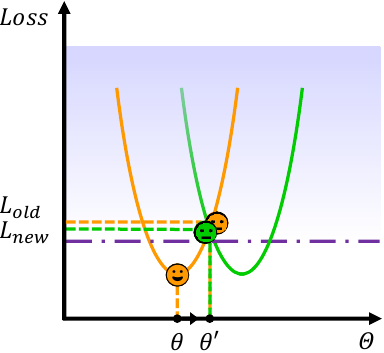}}
    \hfil
    \subfloat[C-Flat]{\includegraphics[width=1.2in]{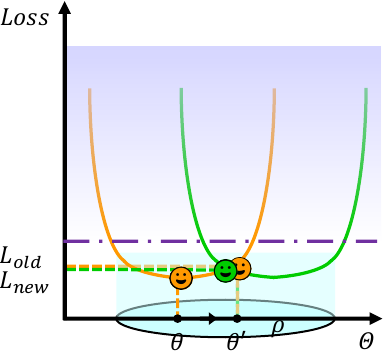}}
    \hfil
    \subfloat{}{\includegraphics[width=0.9in]{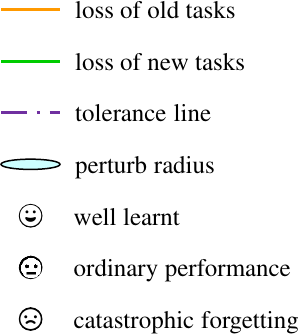}}
    % \subfloat[]{\includegraphics[width=1.in]{main_figs/f4.pdf}%
    %     \label{}}
% \includegraphics[width=0.7\textwidth]{main_figs/framework_v2.pdf}
\caption{Illustration of C-Flat overcoming catastrophe forgetting by fine-tuning the old model parameter to flat minima of new task.
a) loss minima for current task only can cause catastrophe forgetting on previous ones.
b) balanced optima aligned by regularization leads to unsatisfying results for both old and new tasks.
c) C-Flat seeks global optima for all tasks with flattened loss landscape.}
\label{fig:framework}
\vspace{-3mm}
\end{figure*}

\textbf{Our solution.} 
Inspired by these works, a beyond zeroth-order sharpness-aware continual optimization method is proposed, as demonstrated in Figure \ref{fig:traj} and \ref{fig:framework}, where loss landscape flatness is emphasized to enhance the model's generalization ability.
Thus, the model can consistently converge to a flat minimum in each phase, and then smoothly transition to the global optimum of the joint knowledge space of the current and subsequent tasks, thereby mitigating catastrophic forgetting in CL.
We dub this method \textbf{C}ontinual \textbf{Flat}ness (C-Flat or $C\flat$), as introduced in our conference paper. In this work, we extend C-Flat to C-Flat++, which converges faster than the original C-Flat.

\textbf{Contribution.}
We propose C-Flat, a simple and flexible optimization method designed for continual learning. By guiding the model toward flatter regions of the loss landscape, C-Flat improves generalization while maintaining minimal implementation cost. Its seamless integration with various continual learning approaches makes it a practical and versatile solution for enhancing performance across different settings.

We present a unified C-Flat framework, applicable to a wide range of continual learning methods, including regularization-based, memory-based, and expansion-based approaches. This versatility demonstrates C-Flat's role as a general-purpose optimizer. Extensive experiments on multiple benchmarks validate that enforcing flatness consistently leads to improved performance, reinforcing the principle that \textbf{"Flatter is Better"} in nearly all cases. 
% These findings emphasize the importance of loss landscape geometry in tackling the challenges of continual learning.

To the best of our knowledge, this work is the first to comprehensively compare various continual learning approaches with loss landscape-aware optimization, establishing a solid baseline for future research in this area.

\textbf{Difference.} Building upon our prior conference version \cite{bian2025make}, this work introduces four major improvements that substantially enhance the original approach.

i) We conduct a thorough analysis of the underlying mechanism through which C-Flat promotes flatter regions in the loss landscape. Building on these insights, we propose C-Flat++, a refined variant that incorporates selective parameter updates to accelerate convergence.

ii) Experimental results demonstrate that C-Flat++ achieves comparable or superior performance while significantly reducing training overhead—requiring, on average, only ~30\% of the computational cost of the original C-Flat.

iii) We extend the benchmark evaluation to include CUB and Omnibenchmark, two widely used datasets in continual learning \cite{zhou2024continual}, offering a more complete and robust evaluation.

iv) Furthermore, we explore the applicability of C-Flat++ in domain-incremental learning with pre-trained models, further extending its utility across diverse continual learning scenarios.

\section{Related work}

\textbf{Continual learning} methods roughly are categorized into three groups: \textit{Memory-based} methods write experience in memory to alleviate forgetting. Some work~\cite{rebuffi2017icarl, NEURIPS2019_fa7cdfad, jeeveswaran2023birt, sun2023regularizing} design different sampling strategies to establish limited budgets in a memory buffer for rehearsal. However, these methods require access to raw past data, which is discouraged in practice due to privacy concerns. Instead, recently a series of works~\cite{deng2021flattening,lin2022trgp,saha2020gradient, lin2023pcr, sun2023decoupling} elaborately construct special subspace of old tasks as the memory. \textit{Regularization-based} methods aim to realize consolidation of the previous knowledge by introducing additional regularization terms in the loss function. Some works~\cite{DBLP:journals/pami/LiH18a, kirkpatrick2017overcoming, cha2020cpr} enforce the important weights in the parameter space~\cite{rudner2022continual, kim2022warping, akyurek2021subspace}, feature representations~\cite{bhat2023task, gao2022r}, or the logits outputs~\cite{DBLP:journals/pami/LiH18a, oh2022alife} of the current model function to be close to that of the old one. \textit{Expansion-based} methods dedicate different incremental model structures towards each task to minimize forgetting~\cite{zhou2022model, lu2024revisiting}. Some work~\cite{DBLP:conf/icml/SerraSMK18, hu2023dense} exploit modular network architectures (dynamically extending extra components~\cite{DBLP:conf/cvpr/YanX021, zhu2022self}, or freeze partial parameters~\cite{liu2021adaptive}) to overcome forgetting. Trivially, methods in this category implicitly shift the burden of storing numerous raw data into the retention of model~\cite{zhou2022model}.

\textbf{Gradient-based solutions} are a main group in CL, including shaping loss landscape, tempering the tug-of-war of gradient, and other learning dynamics~\cite{hadsell2020embracing, delange2021continual, li2024unigrad, DBLP:journals/jmlr/MehtaPCS23, feng2025zeroflow}. 
One promising solution is to modify the gradients of different tasks and hence overcome forgetting~\cite{chaudhry2018efficient, lopez2017gradient}, e.g., aligning the gradients of current and old one~\cite{farajtabar2020orthogonal, feng2022progressive}, or, learning more efficient in the case of conflicting objectives~\cite{wang2021gradient}. Other solutions~\cite{deng2021flattening, shi2021overcoming} focus on characterizing the generalization from the loss landscape perspectives to improve CL performance and yet are rarely explored.

\textbf{Sharpness minimization in CL} Many recent works~\cite{he2019asymmetric, foret2020sharpness, baldassi2020shaping} are proposed to optimize neural networks in standard training scenarios towards flat minima. 
Wide local minima were considered an important regularization in CL to enforce the similarity of important parameters learned from past tasks~\cite{cha2020cpr}. Sharpness-aware seeking for loss landscape flat minima starts to gain more attention in CL, especially SAM based zeroth order sharpness is well discussed. An investigation \cite{DBLP:journals/jmlr/MehtaPCS23} proves SAM can help with addressing forgetting in CL, and \cite{DBLP:journals/kbs/ChenJWC24} proposed a combined SAM for few-shot CL.
SAM is also used for boosting the performance of specific methods like DFGP~\cite{DBLP:conf/iccv/YangSWLG023} and FS-DGPM~\cite{deng2021flattening} designed for GPM. SAM-CL~\cite{DBLP:conf/soict/TungVHT23} series with loss term gradient alignment for memory-based CL. These efforts kicked off the study of flat minima in CL, however, zeroth-order sharpness may not be enough for flatter optimal~\cite{zhuang2022surrogate}. Thus, flatness with a global optima and universal CL framework is further studied.

% In CL, the flat minima also are proven to be effective for mitigating catastrophic forgetting~\cite{deng2021flattening, shi2021overcoming, unifs}.
% For example, FS-DPGM~\cite{deng2021flattening} proposed a dynamic gradient memory projection method using sharpness evaluation to address the ’sensitivity-stability’ dilemma in pseudo rehearsal based CL. F2M~\cite{shi2021overcoming} introduced a fine-tuning strategy for the model parameters within the flatness minima regions only in the base session for incremental few-shot learning. 
% These efforts kicked off the study of flat minima in CL by proving that sharpness-aware minimization~\cite{DBLP:conf/iclr/ForetKMN21} can improve the performance of incremental class learning. In this work, a beyond zeroth-order flatness and universal CL framework is further studied.

\section{Method}

% In this section, we first present the preliminary concepts related to flat sharpness, which include zeroth-order and first-order sharpness. Then, we introduce the C-Flat solutions for various CL architectures, as detailed in the conference version. Moreover, an adaptive strategy is employed to selectively apply sharpness for accelerating optimization without compromising convergence.            

\subsection{Preliminary}
Our solution addresses the learning sensitivity-stability dilemma in CL by enhancing model generalization for joint learning across knowledge obtained from different catalogs, domains, or tasks. Additionally, we propose a general yet more powerful optimization method, strengthened by the latest gradient landscape flatness, as a 'plug-and-play' tool for any CL approach.

\subsubsection{Loss landscape sharpness}
Let $f$ denote a model with parameter $\theta$ and loss function $ mathcal{L}$. In conventional neural network optimization, the parameter $\theta$ typically converges to a point with low loss along the gradient direction $\nabla \mathcal{L}(\theta)$. For simplicity, we henceforth represent $\mathcal{L}(f(\theta))$ as $\mathcal{L}(\theta)$ in the following sections, unless otherwise specified.
However, this often leads to convergence in a sharp valley of the loss surface, which results in poor generalization. In other words, a small perturbation to the model parameters when adapting to new tasks can cause significant performance degradation and lead to catastrophic forgetting in continual learning. Instead, the zeroth-order sharpness constrains the maximal loss difference within a neighborhood around $\theta$. Let $B(\theta, \rho) = \{\theta' : \|\theta' - \theta\| < \rho\}$ represent the neighborhood of $\theta$ with radius $\rho > 0$ in the Euclidean space $\Theta \subset \mathbb{R}^d$. It can be formulated as:
\begin{equation}
    R^{0}_{\rho}(\theta) = \max \left\{ \mathcal{L}(\theta') - \mathcal{L}(\theta) : \theta' \in B(\theta, \rho) \right\},
\end{equation}
where $\mathcal{L}(\theta)$ denotes the loss of a model with parameter $\theta$ on any dataset, using an oracle loss function $\mathcal{L}(\cdot)$.
The zeroth-order sharpness regularization $R^{0}_{\rho}(\theta)$ can then be directly applied to constrain the maximal loss within the neighborhood:
\begin{equation}
    \mathcal{L}^{R^0_{\rho}}(\theta) = \mathcal{L}(\theta) + R^{0}_{\rho}(\theta) = \max \left\{ \mathcal{L}(\theta') : \theta' \in B(\theta, \rho) \right\}.
\end{equation}

However, for some fixed $\rho$, local minima with lower loss do not always correspond to a smaller maximal Hessian eigenvalue~\cite{zhuang2022surrogate}, which is related to the curvature of the loss surface in the neighborhood. This means that the zeroth-order sharpness optimizer only may converge to a sharper, suboptimal solution rather than to a flatter global optimum with better generalization ability. This issue becomes even more pronounced in continual learning, where the complexity of sophisticated domains, larger models, and multiple optimization objectives can create a loss landscape resembling a jagged terrain with deep, sharp valleys, making optimization even more challenging.

Recently, first-order gradient landscape flatness is proposed as a measurement of the maximal neighborhood gradient norm, which reflects landscape curvature, to better describe the smoothness of the loss landscape:
\begin{equation}
    R^{1}_{\rho}(\theta) = \rho \cdot \max \{ \| \nabla \mathcal{L}(\theta')\|_2 : \theta'\in B(\theta, \rho) \}.
\end{equation}

Unlike zeroth-order sharpness that force the training converging to a local minimal,
first-order flatness alone constraining on the neighborhood smoothness can not lead to an optimal with minimal loss. 

To maximize the generalization ability of loss landscape sharpness for continual learning task, 
we propose a zeroth-first-order sharpness aware optimizer C-Flat for CL.
Considering the data space, model or blocks to be trained are altered regarding the training phase $T$ and CL method, (as detailed in the next subsection), we define the C-Flat loss as follows:
\begin{align}
\label{eq:cf_loss}
\mathcal{L}^{C}(\theta^T) &=\mathcal{L}(\theta^T) + R^0_{\rho}(\theta^T) 
+ \lambda \cdot R^1_{\rho}(\theta^T) \nonumber \\
&= \mathcal{L}^{R^0_{\rho}}(\theta^T) + \lambda \cdot R^1_{\rho}(\theta^T),
\end{align}
with the minimization objective:
\begin{align}
\min_{\theta^T} \underset{\theta^T_0, \theta^T_1}{\max} \left\{ \mathcal{L}(\theta^T_0) + \lambda \rho \cdot \|\nabla \mathcal{L}(\theta^T_1)\|_2 : \theta^T_0, \theta^T_1 \in B(\theta^T, \rho) \right\}, \nonumber
\end{align}
where $\mathcal{L}^{R^0_{\rho}} (\theta)$ is constructed to replace the original CL loss, while $R^{1}_{\rho}(\theta)$ further regularizes the smoothness of the neighborhood, 
and hyperparameter $\lambda$ is to balance the influence of $R^{1}_{\rho}$ as an additional regularization to loss function $\mathcal{L}$.
Hence, the local minima within a flat and smooth neighborhood is calculated for a generalized model possessing both old and new knowledge.

\subsubsection{Optimization}
In our conference work, the two regularization terms in the proposed C-Flat are resolved correspondingly in each iteration.
Assuming the loss function $\mathcal{L}(\cdot)$ is differentiable and bounded, 
the gradient of $\mathcal{L}^{R^0_{\rho}}$ at point $\theta^T$ can be approximated by 
\begin{align}
\label{eq:SAM}
    \nabla \mathcal{L}^{R^0_{\rho}}(\theta^T) \approx \nabla \mathcal{L}(\theta^T_0) 
    \text{\, with } \theta^T_0  = \theta^T + \rho \cdot \frac{\nabla \mathcal{L}(\theta^T)} {\|\nabla \mathcal{L}(\theta^T)\|_2}.
\end{align}
And the gradient of the first-order flatness regularization $\nabla R^{1}_{\rho}(\theta^T)$ can be approximated by
\begin{align}
\label{eq:GAM}
    &\nabla R^{1}_{\rho}(\theta^T)  \approx \rho \cdot \nabla \|\nabla \mathcal{L}(\theta^T_1)\|_2, \\
    \text{with \qquad} &\theta^T_1 = \theta^T + \rho \cdot \frac{\nabla \|\nabla \mathcal{L}(\theta^T)\|_2}{\Large\|\nabla \|\nabla \mathcal{L}(\theta^T)\|_2\Large\|_2}, \nonumber\\
    \text{where \qquad} &\nabla \|\nabla \mathcal{L}(\theta^T)\|_2 
    = \nabla^2 \mathcal{L}(\theta^T) \cdot \frac{\nabla \mathcal{L}(\theta^T)}
    {\|\nabla \mathcal{L}(\theta^T)\|_2} \nonumber.
\end{align}

The optimization is detailed in Algorithm~\ref{alg:opt}.
Note that $\nabla \mathcal{L}$ is the gradient of $\mathcal{L}$ with respect to $\theta$ through this paper, and instead of the expensive computation of Hessian matrix $\nabla^2 \mathcal{L}$, Hessian-vector product calculation is used in our algorithm, where the time and especially space complexity are greatly reduced to $o(n)$ using 1 forward and 1 backward propagation.
Thus, the overall calculation in one iteration takes 2 forward and 4 backward propagation in total.

\begin{algorithm}[t]
\caption{C-Flat Optimization}\label{alg:opt}
\begin{algorithmic}[1]

\Statex \textbf{Input:} Training phase $T$, training data $S^T$, model $f^{T-1}$ with parameter $\theta^{T-1}$ from last phase if $T>1$, batch size $b$, oracle loss function $\mathcal{L}$, learning rate $\eta>0$, neighborhood size $\rho$, trade-off coefficient $\lambda$, small constant $\epsilon$.
\Statex \textbf{Output:} Model trained at the current time $T$ with C-Flat.
\Statex \textbf{Initialization:}{
\If{T=1:} 
 Randomly Initialize parameter $\theta^{T=1}, \eta^{T=1}=\eta, \rho^{T=1}=\rho$.
\Else 
\State Reconstruct the model and training set if necessary, 
\State Initialize model parameter $\theta^T$ according to the training strategy, like randomly initialization or $\theta^T=\theta^{T-1}$ in pre-trained model based approaches, $\eta^{T}=\eta, \rho^{T}=\rho$, 
\State Frozen part of the parameter if required.
\EndIf
}
\Statex \textbf{Optimization:}
\While{$\theta^T$ not converge,} 
    \State Sample batch $B^T$ of $b$ random instances from $S^T$
    \State Compute batch's loss gradient $g_{B^T}=\nabla \mathcal{L}({\theta^T})$
    \State Compute $R^0_{\rho}$ perturbation: 
    $\epsilon_{0} = \rho^T \cdot  \frac{|g|} {\|g\|_{2} + \epsilon }$
    \State Approximate zeroth-order gradient: 
    $g_0 = \nabla \mathcal{L}({\theta^T} + \epsilon_{0})$
    \State Compute hessian vector product:
    $h=\nabla^2 \mathcal{L}(\theta) \cdot \frac{\nabla \mathcal{L}({\theta^T})}{\|\nabla \mathcal{L}({\theta^T})\|_2 + \epsilon }$
    \State Compute $R^1_{\rho}$ perturbation: 
    $\epsilon_1 = \rho^T \cdot \frac{h}{\|h\|_2 + \epsilon }$
    \State Approximate first-order gradient: 
    $g_1 = \nabla^2 \mathcal{L}({\theta^T} + \epsilon_1) \cdot \frac{\nabla \mathcal{L}({\theta^T} + \epsilon_1)}{\|\nabla \mathcal{L}({\theta^T} + \epsilon_1)\|_2 + \epsilon}$ 
    \State Update: Model parameter: $\theta^{T} = \theta^T - \eta^T(g_0+\lambda g_1)$;
    Update training parameters $\eta^T$,  $\rho^T$ according to a scheduler that the values drop with iterations
\EndWhile 

\State \textbf{Post-Processing} on Model and Training data if required.

\Return Model $f^T$ with parameter $\theta^T$
\end{algorithmic}
\end{algorithm}

Although performance improves, the fold-expansion cost in training time may be impractical for time-critical systems. Therefore, we investigate solutions aimed at reducing training time in C-Flat++. In the following section, we tackle this challenge by examining the frequency and magnitude of perturbations.

\subsubsection{Theoretical analysis}
Given $R^{0}_{\rho}(\theta)$ measuring the maximal limit of the training loss difference, the first-order flatness is its upper bound by nature.
Denoting $\theta+\epsilon \in B(\theta, \rho)$ the local maximum point, a constant $\epsilon^* \in [0, \epsilon]$ exists according to the mean value theorem that 
\begin{align}
    R^{0}_{\rho}(\theta)  &=max \{ \mathcal{L}(\theta')-\mathcal{L}(\theta) : \theta'\in B(\theta, \rho) \}\nonumber \\
    &= \mathcal{L}(\theta+\epsilon)-\mathcal{L}(\theta) =  \left(\nabla \mathcal{L}(\theta+\epsilon^*) \right)^T \cdot \epsilon \nonumber \\
    &\le \| \nabla \mathcal{L}(\theta+\epsilon^*)\|_2 \cdot \|\epsilon\|_2  \nonumber\\
    &\le max \{ \| \nabla \mathcal{L}(\theta')\|_2: \theta'\in B(\theta, \rho) \} \cdot \rho = R^{1}_{\rho}(\theta) .
\end{align}

% Assuming the loss function is twice differentiable, bounded by $M$, obeys the triangle inequality,
% its gradient has bounded variance $\sigma^2$, and both the loss function and its second-order gradient are $\beta-$Lipschitz smooth,
% we can prove that, according to \cite{DBLP:conf/icml/AndriushchenkoF22,DBLP:conf/cvpr/ZhangXY0023}, C-Flat converges in all tasks
% with $\eta \leq {1}/{\beta}, \rho \leq {1}/{4\beta}$, and $\eta^T_i = {\eta}/{\sqrt{i}}$, $\rho^T_i = {\rho}/{\sqrt[4]{i}}$ for epoch $i$ in any task $T$, 
% \begin{align}
%     &\frac{1}{n^T}\sum_{i=1}^{n^T} \mathbb{E}[ \|\nabla\mathcal{L}^{C}(\theta^T)\|^2] \nonumber\\
%      &\leq  \frac{2}{n^T}\sum_{i=1}^{n^T} \mathbb{E}[\|\nabla\mathcal{L}^{R^0_{\rho}}(\theta^T)\|^2] 
%     + \frac{2}{n^T}\sum_{i=1}^{n^T} \mathbb{E}[\|\lambda R^1_{\rho,S^T}(\theta^T) \|^2] \nonumber\\
% &\leq  \frac{8M\beta}{\sqrt{n^T}}
% + \frac{16\sigma^2}{3b\sqrt{n^T}}
% + \frac{32\lambda^2(2\sqrt{n^T}-1)}{\beta^2 n^T},
% \end{align}
% where $n^T$ is the total iteration numbers of task $T$, and $b$ is the batch size.
\textbf{Assumptions 1:} the loss function is twice differentiable, bounded by $M$, with bounded variance $\sigma^2$, and obeys the triangle inequality. Both the loss function and its second-order gradient are $\beta-$Lipschitz smooth. $\eta \leq {1}/{\beta}, \rho \leq {1}/{4\beta}$, and $\eta^T_i = {\eta}/{\sqrt{i}}$, $\rho^T_i = {\rho}/{\sqrt[4]{i}}$ for epoch $i$ in any task $T$.

\textbf{Claim 1:} with \textbf{Assumptions 1}, 
the convergency of zeroth-sharpness with batch size $b$ is guaranteed \cite{DBLP:conf/icml/AndriushchenkoF22} by 
\begin{align}
\frac{1}{n}\sum_{i=1}^{n} \mathbb{E}[\|\nabla\mathcal{L}^{R^0_{\rho}}\|^2] 
\leq 
\frac{4\beta}{\sqrt{n^T}}[\mathcal{L}(\theta)-\mathcal{L}(\theta^*)]
+ \frac{8\sigma^2}{3b\sqrt{n^T}},
\end{align}
hence, the zeroth-order part of C-Flat is bounded: 
\begin{align}
\frac{1}{n^T}\sum_{i=1}^{n^T} \mathbb{E}[\|\nabla\mathcal{L}^{R^0_{\rho}}(\theta^T)\|^2] 
&\leq 
\frac{4\beta}{\sqrt{n^T}}[\mathcal{L}(\theta^T)] 
+ \frac{8\sigma^2}{3b\sqrt{n^T}} \nonumber \\
&\leq \frac{4M\beta}{\sqrt{n^T}}
+ \frac{8\sigma^2}{3b\sqrt{n^T}}.
\end{align}

\textbf{Lemma 1:} let $\mathscr{\xi}_{tr}(\theta)=\mathscr{l}_{tr}(f^T(\theta), f^T(\theta^*))$, with \textbf{Assumptions 1}, 
the first-order part is bounded by 
\begin{align}
&\frac{1}{n^T}\sum_{i=1}^{n^T} \mathbb{E}[\|\nabla\mathcal{L}^{R^1_{\rho}}(\theta^T)\|^2] \nonumber\\
&\leq
\frac{1}{n^T}\sum_{i=1}^{n^T} \mathbb{E}[\|\nabla\mathscr{\xi}_{tr}(\theta^T+\epsilon_1) -\nabla\mathscr{\xi}_{tr}(\theta^T))\|^2] \nonumber \\
&\leq \frac{\beta^2}{n^T} \sum_{i=1}^{n^T} \mathbb{E}[\| \epsilon_1 \|^2]
\leq \frac{\beta^2 \eta^2}{n^T} \sum_{i=1}^{n^T} \mathbb{E}[\| \rho_i^T \|^2]  \nonumber\\
&\leq \frac{\rho^2}{n^T} \sum_{i=1}^{n^T} \mathbb{E}[i^{-2}] \leq \frac{16 (2\sqrt{n^T}-1)}{\beta^2 n^T}.
\end{align}

\textbf{Theorem 1:} with \textbf{Assumptions 1}, by combining the zeroth- and first-order parts, we can prove C-Flat converges in all tasks that 
\begin{align}
    &\frac{1}{n^T}\sum_{i=1}^{n^T} \mathbb{E}[ \|\nabla\mathcal{L}^{C}(\theta^T)\|^2] \nonumber\\
     &\leq  \frac{2}{n^T}\sum_{i=1}^{n^T} \mathbb{E}[\|\nabla\mathcal{L}^{R^0_{\rho}}(\theta^T)\|^2] 
    + \frac{2}{n^T}\sum_{i=1}^{n^T} \mathbb{E}[\|\lambda R^1_{\rho,S^T}(\theta^T) \|^2] \nonumber\\
&\leq  \frac{8M\beta}{\sqrt{n^T}}
+ \frac{16\sigma^2}{3b\sqrt{n^T}}
+ \frac{32\lambda^2(2\sqrt{n^T}-1)}{\beta^2 n^T}.
\end{align}
Let $\nabla^2 \mathcal{L}(\theta^*)$ denotes the Hessian matrix at local minimum $\theta^*$,
its maximal eigenvalue $\lambda_{max} (\nabla^2 \mathcal{L}(\theta^*))$ is a proper measure of the landscape curvature.
The first-order flatness is proven to be related to the maximal eigenvalue of the Hessian matrix as $R^{1}_{\rho}(\theta^*) = \rho^2 \cdot \lambda_{max} (\nabla^2 \mathcal{L}(\theta^*))$, thus the C-Flat regularization can also be used as an index of model generalization ability, with the following upper bound:
\begin{equation}\label{eq:Hessian}
    \mathcal{L}^{C}_{\rho}(\theta^*)=R^{0}_{\rho}(\theta^*)+\lambda R^{1}_{\rho}(\theta^*) \leq (1+\lambda)\rho^2 \cdot \lambda_{max} (\nabla^2 \mathcal{L}(\theta^*)).
\end{equation}

\subsection{A Unified CL Framework Using C-Flat}
\label{subsec:CIL}

C-Flat introduces a unified CL framework that accommodates a variety of CL approaches. To maintain focus, this subsection is centered on Class-Incremental Learning (CIL) tasks, which are the most challenging CL scenarios, aiming to develop lifelong learning models for sequentially arriving class-agnostic data. However, the framework is not restricted to CIL; it can be readily extended to other tasks, such as Domain-Incremental Learning (DIL) and Task-Incremental Learning (TIL). We will further explore these extensions in the ablation study.

\subsubsection{Memory-based solutions}
\textbf{Memory-based} methods aim to mitigate catastrophic forgetting by storing samples from previous phases within a memory limit or generating pseudo-samples through generative models to extend the current training data space. In such approaches, memory replay strategies are then employed, where new data from the current task is combined with stored previous samples to preserve the learned representations of seen classes with $\hat{S^T} = S^T \cup Sample^{t<T}$. 

One of the early works in this area is iCaRL, which learns classifiers and feature representations simultaneously while preserving a small set of representative exemplars using Nearest-Mean-of-Exemplars Classification. These exemplars act as a compact memory to retain knowledge from previous tasks. To balance performance across both new and old tasks, iCaRL introduces a loss function: $\mathcal{L}_{\hat{S^T}} = \mathcal{L}^{CE}_{\hat{S^T}} + \mathcal{L}^{KL}_{\hat{S^T}},$ where $\mathcal{L}^{CE}$ is the cross-entropy loss for the current task and $\mathcal{L}^{KL}$ is the knowledge distillation loss to retain knowledge from previous tasks. This combined loss ensures both sensitivity to new tasks and generalization to old ones.

\textbf{Solution:} For memory-based methods, C-Flat can be easily applied to these scenarios by reconstructing the oracle loss function with its zeroth- and first-order flatness measurements, as shown in Equation ~\ref{eq:replay}. The model is then trained with Algorithm~\ref{alg:opt}, using a dataset extended with the previous exemplars.
\begin{align}\label{eq:replay}
\mathcal{L}^{C}_{\hat{S^T}}(\theta^T) = \mathcal{L}^{R^0_{\rho}}_{\hat{S^T}}(\theta^T)
+ \lambda \cdot \mathcal{L}^{R^1_{\rho}}_{\hat{S^T}}(\theta^T) .
\end{align}

\subsubsection{Regularization-based solutions}
\textbf{Regularization-based} approaches aim to apply regularization techniques to the model to preserve the previously learned knowledge. For example, WA introduces weight alignment in the final inference phase to balance the performance between old and new classes. Let $\phi$ represent the feature learning layers of the model, and let $\psi = [\psi^{old}, \psi^{new}]$ denote the decision head for all classes, with two branches corresponding to old and new data classes. The output is adjusted as follows:
$f(x) = [\psi^{old}(\phi(x)), \gamma \cdot \psi^{new}(\phi(x))]$, where $\gamma$ is the ratio of the average norm of $\psi^{old}$ and $\psi^{new}$ over all classes.
PODNet introduces an additional loss term to enforce consistency in the model's embedding layers by regularizing the difference in the output of intermediate convolution layers, $f^T_c$, across both height and width.

Another key group of regularization-based methods is Gradient Projection Memory (GPM), which explicitly aligns the gradient direction with the new knowledge to be learned. GPM stores a minimal set of bases for the Core Gradient Space, referred to as the Gradient Projection Memory. This allows gradient steps to be taken only along the orthogonal direction, helping to learn new features without forgetting the essential knowledge from earlier stages. FS-DGPM improves upon this by updating model parameters along the orthogonal gradient direction at the zeroth-order sharpness minima within the dynamic GPM space.

\textbf{Solution:} For regularization-based methods, the same plug-and-play technique can be applied to reformulate the loss function as in Equation \ref{eq:replay}, and optimization can proceed using Algorithm\ref{alg:opt}.

\textbf{An alternative solution} for the gradient-based methods like GPM and the improved FS-DGPM, is to introduce C-Flat optimization at the gradient alignment stage, so that the orthogonal gradient at a flatter minima is used to ensure that the training can cross over the knowledge gap between different data categories. 
The implementation of our C-Flat-GPM is detailed in Algorithm~\ref{alg:FS}. 
\begin{algorithm}[t]
\caption{C-Flat for GPM-family at $T>1$}\label{alg:FS}
\begin{algorithmic}
\Statex \textbf{Input:} Training set $\hat{S^T}$, parameter $\theta^T=\theta^{T-1}$, loss $\mathcal{L}$, learning rate $\eta_1, \eta_2$, basis matrix $\mathbb{M}$ and significance $\Lambda$ from replay buffer.
\While{$\theta^T$ not converge,}
\State Sample batch $B^T$ from $S^T$
\State Compute perturbation $\epsilon_c$ using C-Flat optimization
\State Update basis significance: $\Lambda=\Lambda - \eta_1\cdot \bigtriangledown_{\Lambda} \mathcal{L}_{B^T}(\theta^T+\epsilon_c)$
\State Update model parameter: $\theta^T = \theta^T - \eta_2 \cdot (I-\mathbb{M}\Lambda M)\bigtriangledown \mathcal{L}_{B^T}(\theta^T+\epsilon_c)$
\State Update $M$ and replay buffer
\EndWhile

\Return Model parameter $\theta^T$
\end{algorithmic}
\end{algorithm}

\subsubsection{Expansion-based solutions}
\textbf{Expansion-based} methods explicitly design task-specific parameters to address the challenge of learning and inferring new classes. For example, the Memory-efficient Expandable Model (Memo) splits the embedding module into deep and shallow layers, such that $\phi = \phi_f(\phi_g)$, where $\phi_f$ and $\phi_g$ represent specialized blocks for different tasks and a generalized block that can be shared across training phases. To handle new classes, an additional block $\phi_f^{new}$ is added to the deep layers for extracting specific features for those classes. The model can then be reformulated as $f^T = \psi^T([\phi_f^{T-1}(\phi_g), \phi_f^{new}(\phi_g)])$. In this setup, training the new model concentrates on the task-specific component, while the shared shallow layers remain frozen. The corresponding loss function is given by $\mathcal{L}{\hat{S^T}}^{Memo} = \mathcal{L}^{CE}{\hat{S^T}}(\psi^T([\phi_f^{T-1}(\phi_g), \phi_f^{new}(\phi_g)])))$.

Foster applies a KL-divergence-based loss to regularize the interaction between the old and new blocks, ensuring stable performance on the previous data. Additionally, it introduces a pruning strategy for redundant parameters and features, maintaining a single backbone model through knowledge distillation. DER adopts a similar framework, further introducing an auxiliary classifier and an associated loss term to help the model learn diverse, discriminative features for novel concepts.

\textbf{Solution:} For expansion-based methods, the plug-and-play strategy remains applicable.
The C-Flat loss can be adjusted to fit the reconstructed model, as shown in Equation \ref{eq:expansion}.
Therefore, C-Flat optimization is performed using Algorithm\ref{alg:opt} in the initial stage, where the newly constructed block is optimized while the generalized blocks remain fixed.
The final model is then obtained after post-processing.
\begin{align}\label{eq:expansion}
&\mathcal{L}^{C}_{\hat{S^T}}(f^T)  =\mathcal{L}^{R^0_{\rho}}_{\hat{S^T}}([\psi^{old}, \psi^{new}] ([\phi_f^{T-1}(\phi_g), \phi_f^{new}(\phi_g)]))  \nonumber \\
&+ \lambda \cdot \mathcal{L}^{R^1_{\rho}}_{\hat{S^T}}([\psi^{old}, \psi^{new}]([\phi_f^{T-1}(\phi_g), \phi_f^{new}(\phi_g)])).
\end{align}

\textbf{To conclude}, C-Flat can be easily applied to any CL method with reconstructed loss function, and thus trained with the corresponding optimize as shown in Algorithm~\ref{alg:opt}.
Dedicated design using C-Flat like for the GPM family is also possible wherever flat minima is required.

\subsection{Efficient C-Flat++}
\label{subsec:cflatpp}
\begin{table}
\centering
\caption{Hybrid optimization strategy integrating \textcolor{SGDColor}{SGD} and \textcolor{CFlatColor}{C-Flat}. ''$p\%$'': substitute $p\%$ SGD steps with C-Flat updates. "$\Rightarrow$": initial update sequence: $p\%$ C-Flat + $(1-p\%)$ SGD.
"$\Leftarrow$": reverse update sequence: $(1-p\%)$ SGD + $p\%$ C-Flat.}
\label{tab:subcflat}
\vspace{-1mm}
\begin{tblr}{
  column{2-7} = {c},
  cell{1}{3} = {c=3}{},
  hline{1-3,8} = {-}{},
}
Optimizer& \textcolor{SGDColor}{SGD}   & \textcolor{CFlatColor!50!SGDColor}{Hybrid} &       &       & \textcolor{CFlatColor}{C-Flat} & \textcolor{CFlatColor!35!SGDColor}{C-Flat++}     \\
Proportion & \textcolor{SGDColor}{0\%} & \textcolor{CFlatColor!20!SGDColor}{25\%} & \textcolor{CFlatColor!50!SGDColor}{50\%} & \textcolor{CFlatColor!75!SGDColor}{75\%} &  \textcolor{CFlatColor}{100\%} & \textcolor{CFlatColor!25!SGDColor}{25.1\%}  \\
Img/s               & 2716.9 & 1776.1  & 1304.8 & 1059.7 & 938.5   & 1765.6 \\
Avg ($\Rightarrow$) & 69.13  & 69.43   & 69.91  & 70.15  & 70.79   & 70.39  \\ 
Avg ($\Leftarrow$)  & 69.13  & 69.91   & 70.21  & 70.53  & 70.79   & 70.39  \\  
Last ($\Rightarrow$)& 57.40  & 57.59   & 59.01  & 59.84  & 60.20   & 60.02  \\
Last ($\Leftarrow$) & 57.40  & 58.64   & 59.13  & 59.85  & 60.20   & 60.02       
\end{tblr}
\end{table}

\begin{figure}
\vspace{-5mm}
\centering
    \subfloat[Squared gradient norm\label{fig:snorm_a}]{\includegraphics[width=1.7in]{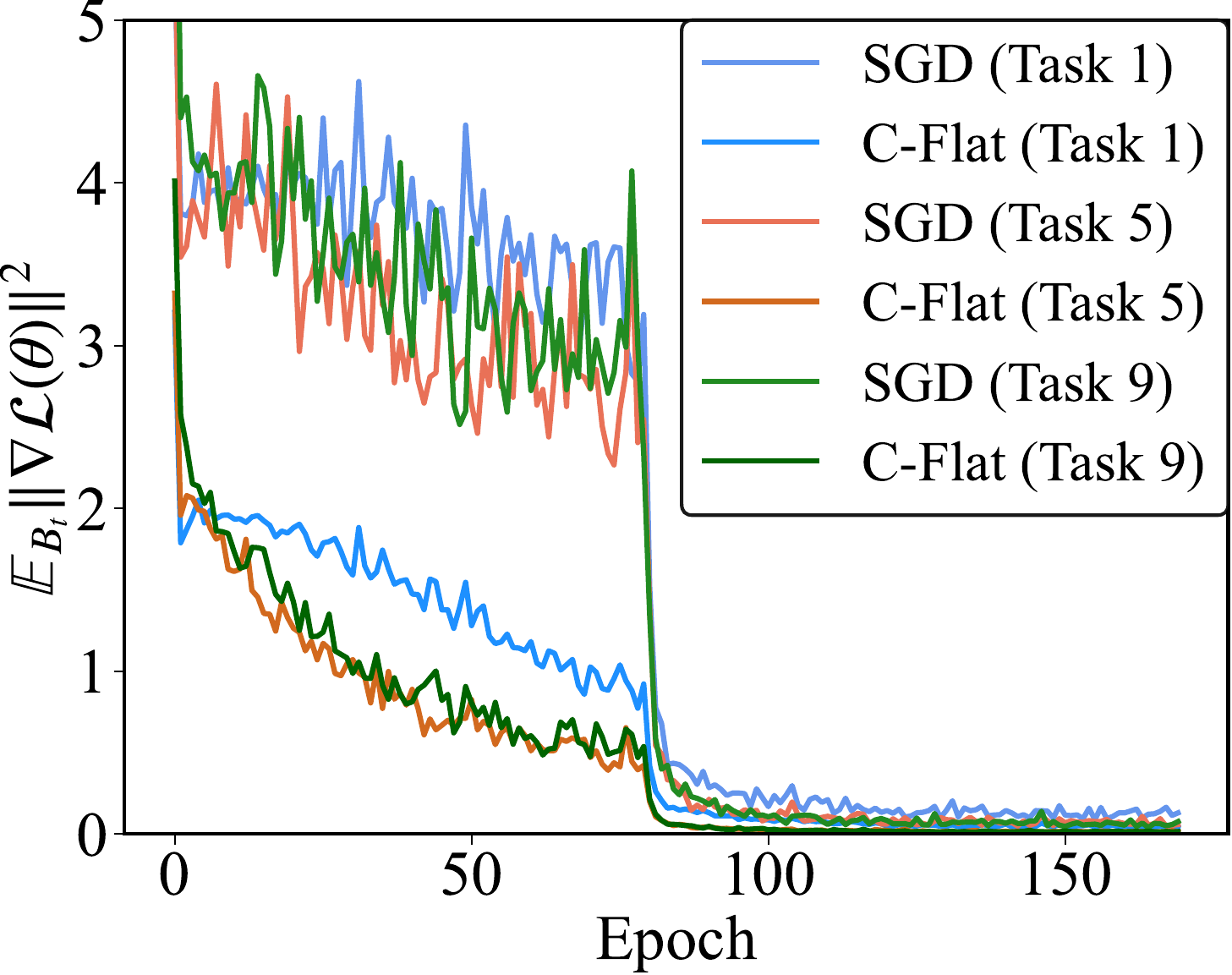}}
    \hfil
    \subfloat[Sharpness proxy\label{fig:snorm_b}]{\includegraphics[width=1.7in]{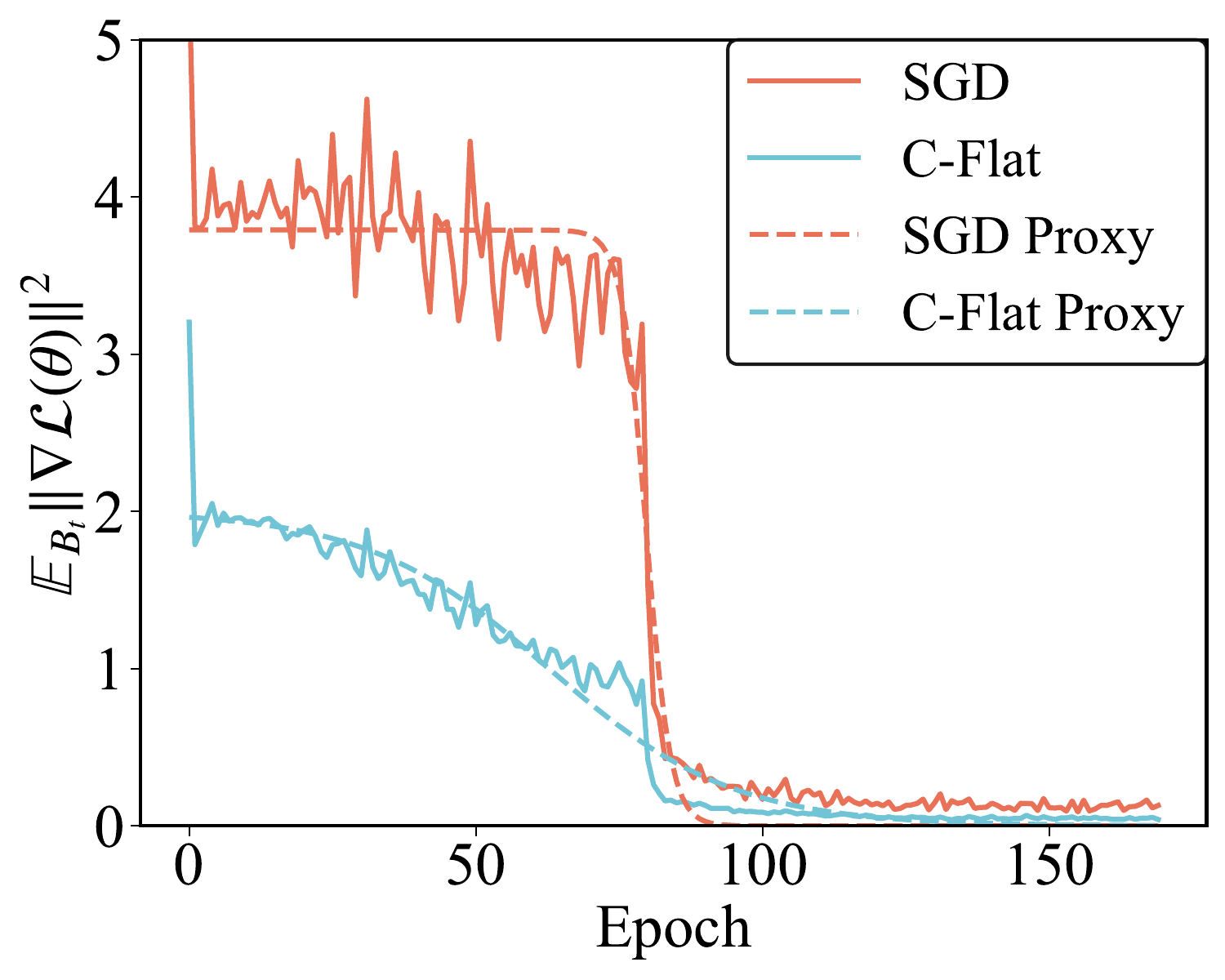}}
\caption{The squared gradient norm $\mathbb{E}_{B_t}\|\nabla\mathcal{L}(\theta)\|^2$ and sharpness proxy of SGD and C-Flat across tasks.}
\label{fig:snorm}
\vspace{-2mm}
\end{figure}

C-Flat modifies the update mechanism of base optimizers by incorporating sharpness-aware computations within each iteration. While demonstrating performance enhancements, this approach introduces significant computational overhead compared to conventional optimization methods. A promising acceleration strategy involves partial substitution of C-Flat with SGD updates. Table \ref {tab:subcflat} presents empirical results of hybrid optimization configurations combining C-Flat with SGD at varying mixture ratios.
Experimental evidence indicates that C-Flat maintains compatibility with SGD updates, where partial sharpness-aware regularization suffices for identifying flat regions in the feature space. However, the hybrid optimization strategy shows limitations in adapting to sophisticated scenarios with complex configuration requirements. To address this, implementing a proxy for assessing local sharpness in the loss landscape is necessary, with selective application of C-Flat only when significant sharpness values are detected. As illustrated in Figure \ref{fig:traj}, this dynamic switching strategy effectively balances computational efficiency with the pursuit of flat minima. 

Standard sharpness evaluation relies on Hessian eigenvalue analysis, which incurs high computational complexity due to the necessity of computing second-order derivatives of the loss function. To address this efficiency constraint, an established approximation method formulates sharpness estimation as $\mathbb{E}_{B_t}\|\nabla \mathcal{L}(\theta)\|^2$, where the gradient $\nabla \mathcal{L}(\theta)$ is derived from stochastically sampled mini-batches $B_t \sim \mathbb{D}$. This indicator leverages gradient information to construct computationally tractable sharpness metrics while maintaining statistical relevance to the underlying data distribution.
A lower value of $\mathbb{E}_{B_t}\|\nabla \mathcal{L}(\theta)\|^2$ indicates reduced gradient variance, $\operatorname{Var}_{B_t}(\nabla \mathcal{L}(\theta))$, which corresponds to a flatter, more stable region of the loss landscape. This flatter region is less sensitive to perturbations, thereby enhancing the model's robustness and generalization capacity. Figure \ref{fig:snorm_a} illustrates the evolution of the squared gradient norm $\mathbb{E}_{B_t}\|\nabla\mathcal{L}(\theta)\|^2$ during training with either SGD or C-Flat. Three critical observations emerge:
i) Both SGD and C-Flat reduce the squared gradient norm as training progresses. Upon convergence to a local minimum, the gradient norm reaches a low value, aligning with the asymptotic convergence property of gradient descent.
ii) C-Flat achieves significantly lower terminal norms (e.g., dark vs. light blue curves for Task 1), suggesting convergence to flatter minima that potentially enhance generalization. Besides, C-Flat demonstrates smoother gradient norm trajectories (e.g., reduced oscillations in Task 5 orange vs. brown curves), attributable to its explicit sharpness regularization in Equation \ref{eq:cf_loss}. 
iii) Post-initial tasks (Task 1), gradient norms progressively converge across later tasks (Task 5/9 curves narrowing at 150 epochs). This reflects \textit{representation stiffness} in shared feature layers: $\frac{\partial \mathcal{L}(\theta)}{\partial \phi_j} \approx 0 \quad \text{for } j > k \text{ (frozen earlier task features)}$. Similar phenomenon also observed in EWC~\cite{kirkpatrick2017overcoming} and MAS~\cite{aljundi2018memory}, suggesting architectural limitations in disentangling task-specific features during sequential learning.

\begin{algorithm}[t]
\caption{C-Flat++ Optimization}\label{alg:cflatpp}
\begin{algorithmic}
\Statex \textbf{Input:} Training set $S^T$, parameters $\theta^T = \theta^{T-1}$, learning rate $\eta^T$, sharpness bound $A$, iteration or epoch $i$, sharpness curvature $k$, stabilization point $i_0$, efficient learning rate $\eta_0$.
\While{$\theta^T$ not converge, i += 1}
\State Sample batch $B^T$ from $S^T$
\State Compute batch's loss gradient $g=\nabla \mathcal{L}({\theta^T})$
\State Compute squared gradient norm $\mathbb{E}_{B_t}\|\nabla\mathcal{L}(\theta)\|^2$
\State Compute sharpness proxy $A / (1 + e^{-k(i - i_0)})$
\State Compute error feedback $E = A / (1 + e^{-k(i - i_0)}) - \mathbb{E}_{B_t}\|\nabla\mathcal{L}(\theta)\|^2$
\State Update sharpness proxy $A = A - \eta_0 E$
\If{$E \leq 0$}
    \State Compute perturbation $\epsilon_c$ using C-Flat optimization
    \State batch gradient $g = g_0+\lambda g_1$
\EndIf
\State Update: Model parameter: $\theta^{T} = \theta^T - \eta^T g$
\EndWhile

\Return Model parameter $\theta^T$
\end{algorithmic}
\end{algorithm}

The squared gradient norm introduces minimal computational overhead when integrated with baseline methods, making it an efficient metric for tracking loss landscape sharpness during parameter updates. As shown in Figure~\ref{fig:snorm_b}, the sharpness proxy follows a sigmoidal trajectory, which can be modeled as $A/(1 + e^{-k(i - i_0)})$, where $A$ denotes the asymptotic sharpness bound, $k$ controls the curvature, and $i_0$ marks the inflection point of stabilization. These hyperparameters can be dynamically adjusted through error feedback based on deviations from instantaneous gradient norms. In the experiments, we adopt a linear updation for the sharpness bound $A$ based on batch-level gradient norms for simplicity, while keeping other parameters fixed. Details are presented in Algorithm \ref{alg:cflatpp}.

\section{Analysis}
\subsection{Experimental Setup}

\textbf{Datasets.} We evaluate the performance on CIFAR-100, ImageNet-100, Tiny-ImageNet, CUB, and Omnibenchmark, which are classical datasets in CL community. Following ~\cite{zhou2023pycil, zhou2023class}, we fix the random seed for class-order shuffling to 1993. Subsequently, we follow two typical class splits in CIL: (i) Divide all $\left\| Y_{b} \right\|$ classes equally into $B$ phases, denoted as \textbf{B0\_Inc}$y$; (ii) Treat half of the total classes as initial phases, followed by an equally division of the remaining classes into incremental phases, denoted as \textbf{B50\_Inc}$y$. In both settings, $y$ denotes the number of new classes learned per task. All experimental results are averaged over 3 to 5 random seeds.

% Replay~\cite{NEURIPS2019_fa7cdfad}
% iCaRL~\cite{rebuffi2017icarl}
% WA~\cite{zhao2020maintaining}
% PODNet~\cite{DBLP:conf/eccv/DouillardCORV20}
% DER~\cite{DBLP:conf/cvpr/YanX021}
% FOSTER~\cite{wang2022foster}
% MEMO~\cite{zhou2022model}

\begin{table*}
\centering
\scriptsize
\caption{Average accuracy (mean $\pm$ std) across all phases using 7 state-of-art methods (span all sorts of CL) w/ and w/o C-Flat and C-Flat++ plugged in. \textit{Average Return} and \textit{Maximum Return} in the last two rows represent the average and maximum boost of C-Flat / C-Flat++ over all methods in each column. Proportion denotes the averaged i\% C-Flat calculation of all tasks used.}
\label{tbl_stronger}
\begin{tblr}{
  colspec={l c ccc cc c c c c},
  hline{1,3,24,26} = {-}{},
  cell{1}{1} = {r=2}{l},
  cell{1}{2} = {r=2}{c},
  cell{1}{3} = {c=3}{c},
  cell{1}{6} = {c=2}{c},
  cell{1}{11} = {r=2}{c},
  cell{3,6,9,12,15,18,21}{2} = {r=3}{c},
}
Method            & Technology & CIFAR-100 &           &           & ImageNet-100 &            & Tiny-ImageNet & CUB    & Omnibenchmark & Proportion         \\
                  &       & B0\_Inc5  & B0\_Inc10 & B0\_Inc20 & B50\_Inc10   & B50\_Inc25 & B0\_Inc40     & B0\_Inc20 & B0\_Inc30 & \\
\hline
Replay                &\textcolor{blue}{Mem.}                 & 58.57$_{\pm 0.20}$ &       58.09$_{\pm 0.62}$ & 59.78$_{\pm 0.35}$ & 62.77$_{\pm 1.87}$ & 71.22$_{\pm 0.38}$ & 48.23$_{\pm 0.34}$ & 37.15$_{\pm 0.25}$ & 57.56$_{\pm 0.64}$ & 0.0\% \\
\textit{w/ C-Flat}        &                                              & 60.65$_{\pm 0.54}$ & 59.42$_{\pm 0.75}$ & 61.05$_{\pm 0.64}$ & 66.00$_{\pm 0.71}$ & 72.94$_{\pm 0.37}$ & 50.71$_{\pm 0.07}$ & 39.72$_{\pm 0.03}$ & 60.31$_{\pm 0.62}$ & 100.0\% \\
\textit{w/ C-Flat++}      &                                              & 60.05$_{\pm 0.21}$ & 59.45$_{\pm 0.84}$ & 60.69$_{\pm 0.43}$ & 65.58$_{\pm 0.49}$ & 73.31$_{\pm 0.56}$ & 50.09$_{\pm 0.20}$ & 39.80$_{\pm 0.35}$ & 59.22$_{\pm 0.43}$ & 27.6\% \\
\hline
iCaRL                & \textcolor{blue}{Mem.}                 & 59.36$_{\pm 0.30}$ &            58.80$_{\pm 1.04}$ & 62.18$_{\pm 0.06}$ & 66.32$_{\pm 0.45}$ & 75.13$_{\pm 0.24}$ & 50.81$_{\pm 0.27}$ & 37.14$_{\pm 0.25}$ & 58.55$_{\pm 0.65}$ & 0.0\% \\
\textit{w/ C-Flat}        &                                              & 59.99$_{\pm 0.18}$ & 59.75$_{\pm 0.55}$ & 62.47$_{\pm 0.18}$ & 65.33$_{\pm 0.79}$ & 75.60$_{\pm 1.06}$ & 52.28$_{\pm 0.24}$ & 39.80$_{\pm 0.07}$ & 60.35$_{\pm 0.42}$ & 100.0\% \\
\textit{w/ C-Flat++}      &                                              & 60.08$_{\pm 0.05}$ & 59.55$_{\pm 0.93}$ & 62.35$_{\pm 0.22}$ & 65.14$_{\pm 0.84}$ & 76.48$_{\pm 0.14}$ & 52.00$_{\pm 0.27}$ & 39.08$_{\pm 0.38}$ & 59.85$_{\pm 0.55}$ & 30.8\% \\
\hline
WA                  & \textcolor{red}{Reg.}             & 65.34$_{\pm 0.25}$ &                  65.63$_{\pm 0.91}$ & 67.54$_{\pm 0.26}$ & 75.07$_{\pm 1.16}$ & 82.31$_{\pm 0.16}$ & 57.59$_{\pm 0.09}$ & 40.94$_{\pm 0.40}$ & 62.14$_{\pm 0.61}$ & 0.0\% \\
\textit{w/ C-Flat}        &                                              & 66.45$_{\pm 0.52}$ & 67.00$_{\pm 0.93}$ & 68.93$_{\pm 0.34}$ & 74.03$_{\pm 0.75}$ & 82.39$_{\pm 0.83}$ & 58.93$_{\pm 0.14}$ & 41.79$_{\pm 0.45}$ & 64.43$_{\pm 0.85}$ & 100.0\% \\
\textit{w/ C-Flat++}      &                                              & 65.73$_{\pm 0.36}$ & 66.89$_{\pm 0.99}$ & 68.59$_{\pm 0.32}$ & 74.74$_{\pm 0.43}$ & 83.16$_{\pm 0.27}$ & 58.60$_{\pm 0.11}$ & 42.07$_{\pm 0.39}$ & 64.00$_{\pm 1.07}$ & 24.6\% \\
\hline
PODNet  & \textcolor{red}{Reg.}/\textcolor{blue}{Mem.}                 & 48.48$_{\pm 0.39}$ & 55.60$_{\pm 0.25}$ & 63.63$_{\pm 0.39}$ & 83.51$_{\pm 0.16}$ & 86.16$_{\pm 0.02}$ & 56.75$_{\pm 0.38}$ & 42.40$_{\pm 0.44}$ & 58.77$_{\pm 0.30}$ & 0.0\% \\
\textit{w/ C-Flat }       &                                              & 50.30$_{\pm 0.41}$ & 57.00$_{\pm 0.48}$ & 63.89$_{\pm 0.39}$ & 84.11$_{\pm 0.12}$ & 86.51$_{\pm 0.07}$ & 57.68$_{\pm 0.33}$ & 42.31$_{\pm 0.36}$ & 60.33$_{\pm 0.27}$ & 100.0\% \\
\textit{w/ C-Flat++}      &                                              & 49.56$_{\pm 0.35}$ & 56.49$_{\pm 0.44}$ & 63.86$_{\pm 0.13}$ & 84.00$_{\pm 0.05}$ & 86.39$_{\pm 0.08}$ & 57.18$_{\pm 0.32}$ & 42.10$_{\pm 0.43}$ & 59.72$_{\pm 0.45}$ & 29.8\% \\
\hline
DER         & \textcolor{orange}{Exp.}                             & 69.45$_{\pm 1.09}$ & 70.13$_{\pm 1.40}$ & 71.19$_{\pm 0.13}$ & 85.56$_{\pm 0.29}$ & 87.57$_{\pm 0.14}$ & 60.41$_{\pm 0.67}$ & 45.26$_{\pm 0.58}$ & 68.05$_{\pm 0.13}$ & 0.0\% \\
\textit{w/ C-Flat}        &                                              & 71.37$_{\pm 0.75}$ & 71.35$_{\pm 1.17}$ & 72.19$_{\pm 0.28}$ & 86.60$_{\pm 0.17}$ & 88.07$_{\pm 0.13}$ & 61.82$_{\pm 0.57}$ & 47.35$_{\pm 0.70}$ & 69.71$_{\pm 0.21}$ & 100.0\% \\
\textit{w/ C-Flat++}      &                                              & 70.68$_{\pm 0.65}$ & 71.03$_{\pm 1.20}$ & 71.92$_{\pm 0.31}$ & 86.52$_{\pm 0.14}$ & 88.07$_{\pm 0.13}$ & 61.52$_{\pm 0.42}$ & 46.54$_{\pm 0.72}$ & 69.45$_{\pm 0.24}$ & 28.9\% \\
\hline
FOSTER         & \textcolor{red}{Reg.}/\textcolor{orange}{Exp.}        & 62.83$_{\pm 0.63}$ & 65.84$_{\pm 1.43}$ & 69.33$_{\pm 0.14}$ & 82.67$_{\pm 0.13}$ & 86.50$_{\pm 0.08}$ & 60.75$_{\pm 0.78}$ & 31.84$_{\pm 1.44}$ & 67.09$_{\pm 0.37}$ & 0.0\% \\
\textit{w/ C-Flat}        &                                              & 63.75$_{\pm 0.40}$ & 67.27$_{\pm 1.49}$ & 70.44$_{\pm 0.26}$ & 85.89$_{\pm 0.12}$ & 87.87$_{\pm 0.10}$ & 62.18$_{\pm 0.35}$ & 29.28$_{\pm 0.83}$ & 68.38$_{\pm 0.57}$ & 100.0\% \\
\textit{w/ C-Flat++}      &                                              & 63.37$_{\pm 0.93}$ & 67.97$_{\pm 0.39}$ & 70.55$_{\pm 0.13}$ & 84.96$_{\pm 0.58}$ & 87.68$_{\pm 0.06}$ & 61.37$_{\pm 0.51}$ & 29.79$_{\pm 0.82}$ & 68.23$_{\pm 0.52}$ & 23.7\% \\
\hline
MEMO            & \textcolor{orange}{Exp.}                          & 67.67$_{\pm 0.78}$ & 68.49$_{\pm 1.74}$ & 69.52$_{\pm 0.92}$ & 75.51$_{\pm 1.43}$ & 82.71$_{\pm 0.13}$ & 60.44$_{\pm 0.42}$ & 44.19$_{\pm 0.82}$ & 65.14$_{\pm 0.35}$ & 0.0\% \\
\textit{w/ C-Flat}        &                                              & 68.91$_{\pm 0.61}$ & 69.48$_{\pm 1.25}$ & 70.32$_{\pm 0.60}$ & 79.13$_{\pm 0.30}$ & 83.27$_{\pm 0.26}$ & 61.67$_{\pm 0.49}$ & 45.56$_{\pm 0.82}$ & 66.91$_{\pm 0.42}$ & 100.0\% \\
\textit{w/ C-Flat++}      &                                              & 68.33$_{\pm 0.55}$ & 69.00$_{\pm 1.39}$ & 70.08$_{\pm 0.35}$ & 78.82$_{\pm 0.19}$ & 83.47$_{\pm 0.21}$ & 61.16$_{\pm 0.47}$ & 45.34$_{\pm 0.78}$ & 66.43$_{\pm 0.17}$ & 25.1\% \\
\hline
\textit{Average Return}   & -     & \textcolor{red}{+1.39}/\textcolor{red}{+0.87}     & \textcolor{red}{+1.24}/\textcolor{red}{+1.11}     & \textcolor{red}{+0.87}/\textcolor{red}{+0.70}     & \textcolor{red}{+1.38}/\textcolor{red}{+1.19}        & \textcolor{red}{+0.72}/\textcolor{red}{+1.00}      & \textcolor{red}{+1.47}/\textcolor{red}{+0.99}         & \textcolor{red}{+0.98}/\textcolor{red}{+0.83} & \textcolor{red}{+1.88}/\textcolor{red}{+1.37}    & - \\
\textit{Maximum Return}   & -     & \textcolor{red}{+2.08}/\textcolor{red}{+1.48}     & \textcolor{red}{+1.43}/\textcolor{red}{+2.13}     & \textcolor{red}{+1.39}/\textcolor{red}{+1.22}     & \textcolor{red}{+3.62}/\textcolor{red}{+3.31}        & \textcolor{red}{+1.72}/\textcolor{red}{+2.10}      & \textcolor{red}{+2.48}/\textcolor{red}{+1.86}         & \textcolor{red}{+2.66}/\textcolor{red}{+2.65} & \textcolor{red}{+2.75}/\textcolor{red}{+1.86}    & - \\
\end{tblr}
\end{table*}

\textbf{Baselines.} To evaluate the efficacy of our method, we integrate it into seven top-performing baselines across different CL categories. Replay~\cite{NEURIPS2019_fa7cdfad} and iCaRL~\cite{rebuffi2017icarl} are classical replay-based methods that use raw data as memory cells. PODNet~\cite{DBLP:conf/eccv/DouillardCORV20} is similar to iCaRL but incorporates knowledge distillation to constrain the logits of pooled representations. WA~\cite{zhao2020maintaining} mitigates prediction bias by regularizing discrimination and fairness. DER~\cite{DBLP:conf/cvpr/YanX021}, FOSTER~\cite{wang2022foster}, and MEMO~\cite{zhou2022model} are network expansion methods that dedicate modular architectures to each task by extending sub-networks or freezing partial parameters. The aforementioned methods span three CL categories~\cite{delange2021continual, van2022three}: \textcolor{blue}{memory-based}, \textcolor{red}{regularization-based}, and \textcolor{orange}{expansion-based} methods.

\textbf{Network and training details.} For a given dataset, we evaluate all methods using the same network architecture: ResNet-32 for CIFAR and Tiny-ImageNet, and ResNet-18 for ImageNet-100, CUB, and Omnibenchmark. All methods are allowed to use memory for previously seen classes, and the total memory usage is determined by the total number of classes multiplied by the allowed memory per class (20, 20, 10, 10, and 20 for the five datasets mentioned above). Only for the latter three datasets, the memory per class is fixed across learning. For fairness and adaptability, all $\rho$ and $\lambda$ values used in C-Flat are fixed at 0.2, while $A, k, i_0, \eta_0$ used in C-Flat++ are set to 5, 0.01, 80, 5e-3, respectively. Unless otherwise specified, the hyperparameters for all models follow the settings in our open-source code \cite{bian2025make}.

\textbf{Evaluation metrics.} We use the average accuracy and last accuracy for evaluation, as proposed in \cite{bian2025make, zhou2023pycil}. The last accuracy refers to the accuracy on the test dataset of all previously seen tasks in the last incremental task, while the average accuracy is the average of all last accuracies over incremental tasks. Img/s denotes the number of images processed during training per second.

\subsection{Make Continual Learning Stronger}

\begin{figure*}[]
\vspace{-3mm}
\centering
    \subfloat[Replay w/ C-Flat++]{\includegraphics[width=1.6in]{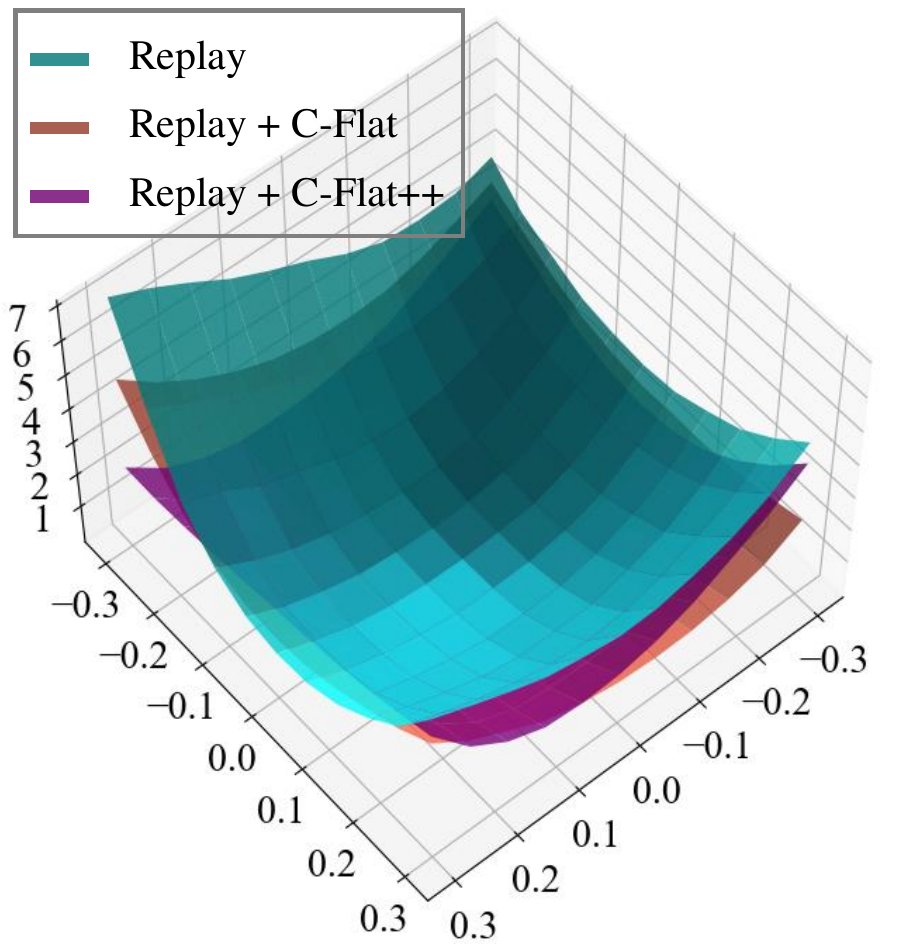}}
    \hfil
    \subfloat[WA w/ C-Flat++]{\includegraphics[width=1.6in]{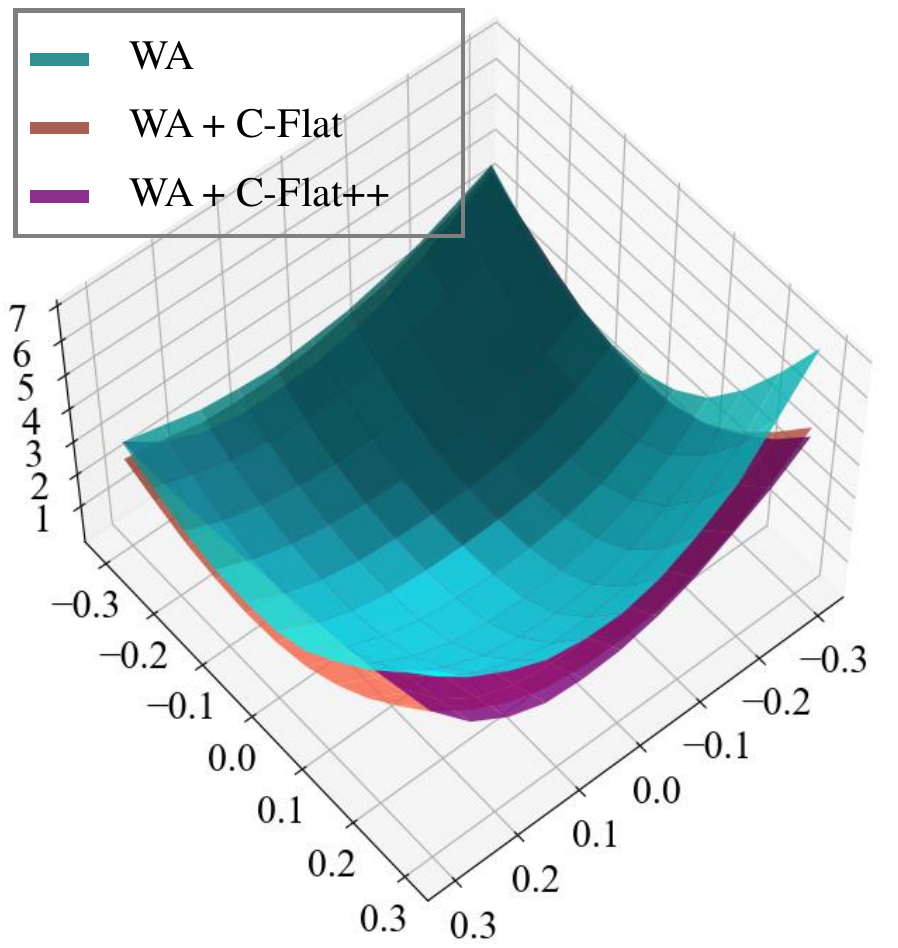}}
    \hfil
    \subfloat[MEMO w/ C-Flat++]{\includegraphics[width=1.6in]{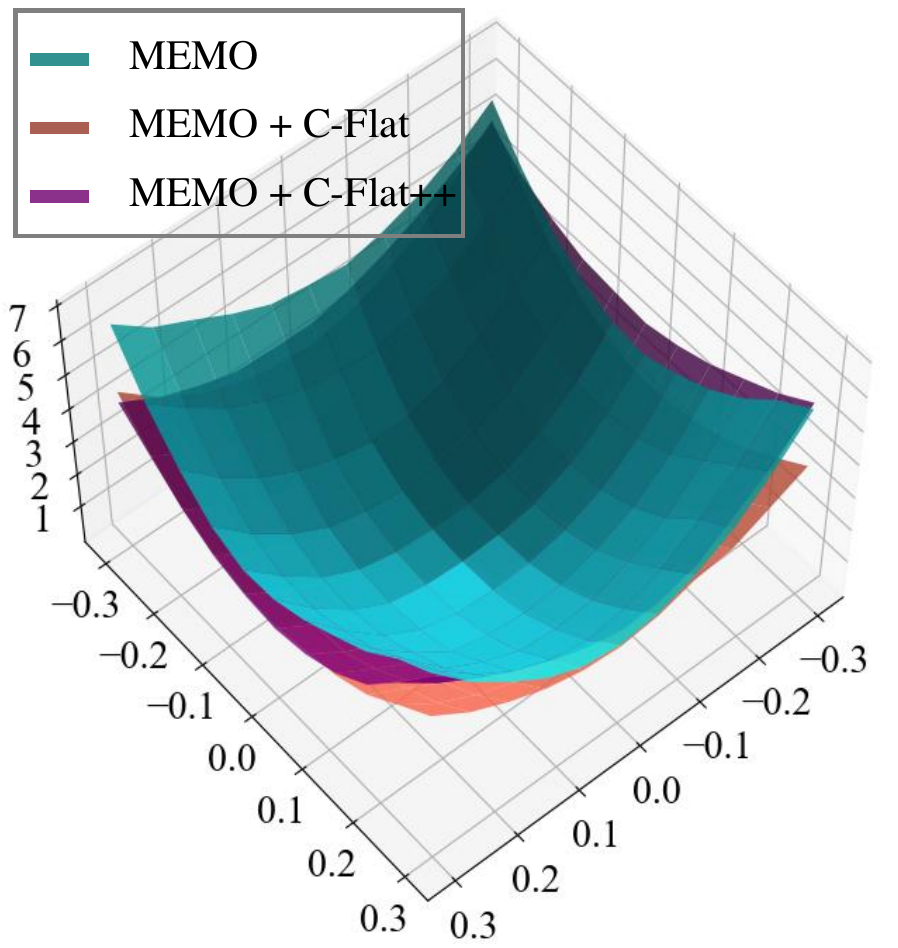}}
\caption{The parametric loss landscapes of Replay (\textcolor{blue}{Mem.}), WA (\textcolor{red}{Reg.}) and MEMO (\textcolor{orange}{Exp.}) are plotted by perturbing the model parameters at the end of training (CIFAR-100, B0\_Inc10) across the first two Hessian eigenvectors.}
\label{fig_loss_landscape}
\vspace{-2mm}
\end{figure*}

\begin{figure}[!t]
\vspace{-3mm}
\centering
    \subfloat[Last accuracy\label{subfig:cflatpp_acc}]{\includegraphics[width=1.7in]{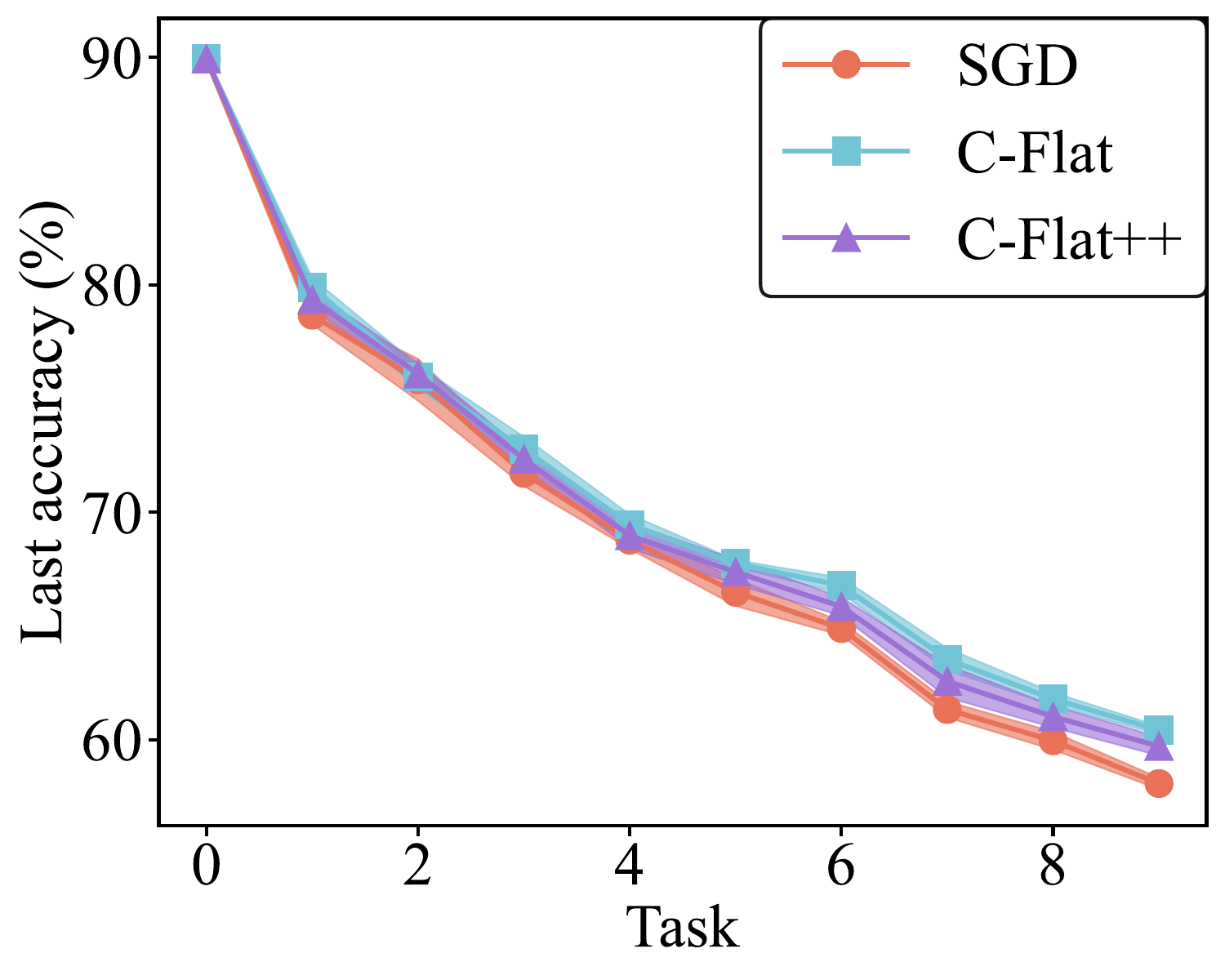}}
    % \hfil
    \subfloat[Squared gradient norm\label{subfig:cflatpp_snorm}]{\includegraphics[width=1.7in]{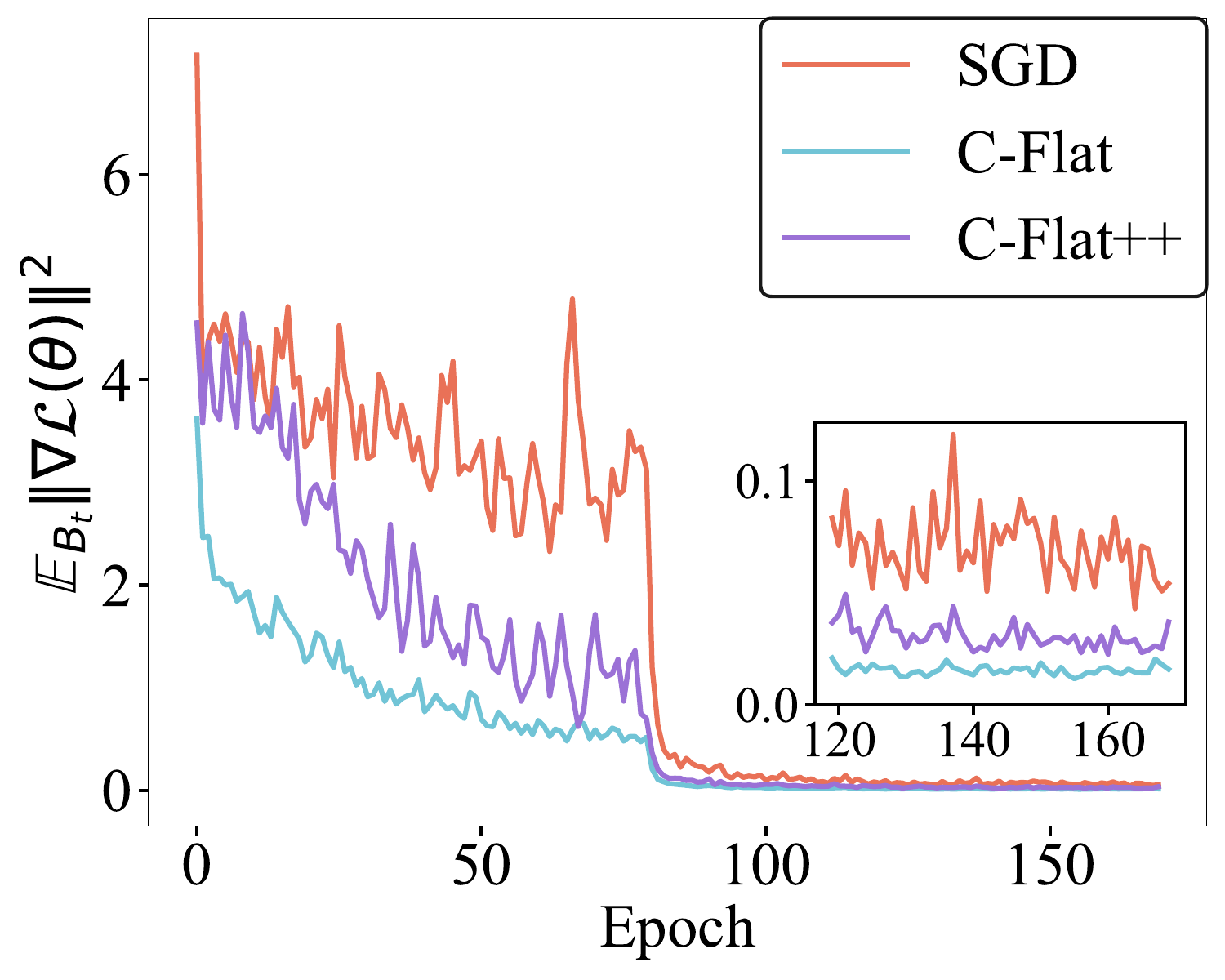}}
    \\
    \subfloat[Running speed\label{subfig:cflatpp_speed}]{\includegraphics[width=1.7in]{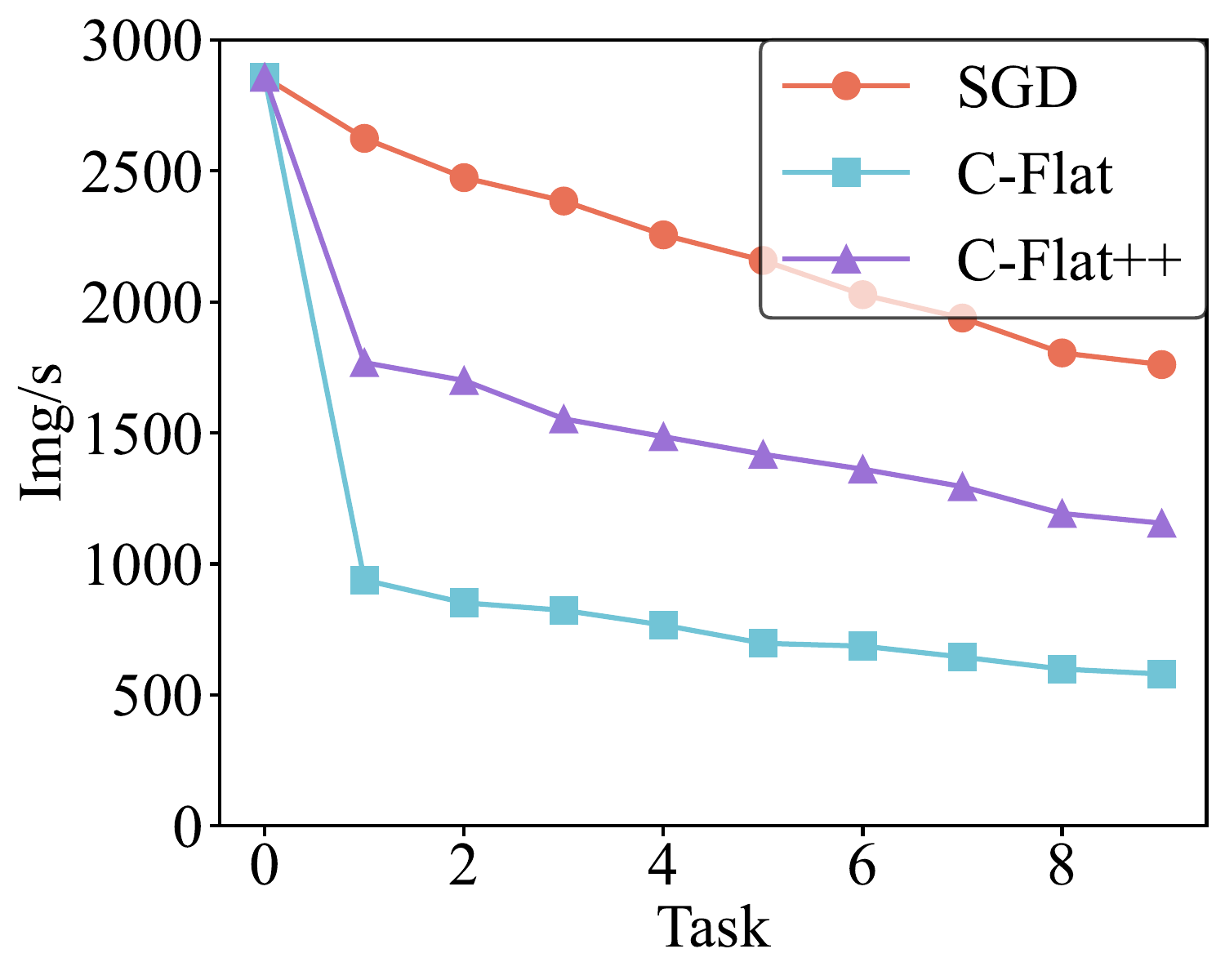}}
    % \hfil
    \subfloat[C-Flat Proportion\label{subfig:cflatpp_prop}]{\includegraphics[width=1.7in]{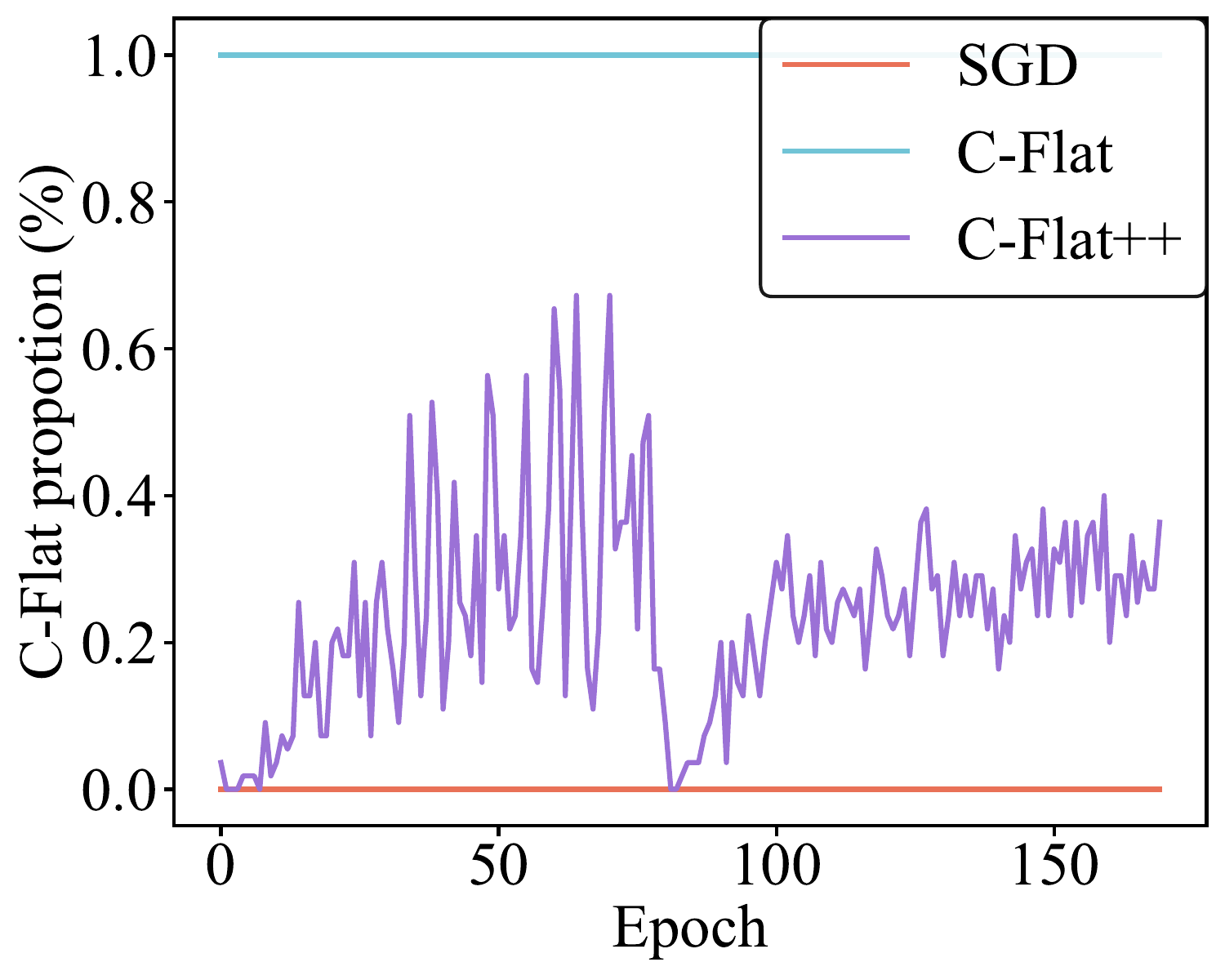}}
    \hfil
  \caption{Analysis of C-Flat++: C-Flat++ with adaptive sharpness-aware minimization performs comparably to C-Flat, while achieving significantly faster acceleration.}
\label{fig:cflatpp}
\end{figure}

Table~\ref{tbl_stronger} empirically demonstrates the superiority of our method: Makes Continual Learning Stronger. In this experiment, we integrate C-Flat and C-Flat++ into seven state-of-the-art methods spanning the full spectrum of CL approaches. From Table~\ref{tbl_stronger}, we observe that
(i) C-Flat consistently enhances performance across all models, covering Memory-based, Regularization-based, and Expansion-based methods. This highlights its plug-and-play nature, enabling seamless integration with various CL paradigms. 
(ii) Across multiple benchmark datasets, including CIFAR-100, ImageNet-100, Tiny-ImageNet, and CUB, C-Flat and C-Flat++ consistently yield improvements, underscoring their generalization capability and adaptability to diverse data distributions. 
(iii) The performance gains persist across multiple incremental learning scenarios, including B0\_Inc5, B0\_Inc10, B0\_Inc20, B50\_Inc10, B50\_Inc25, and B0\_Inc40, reaffirming the robustness of C-Flat and C-Flat++ across different CL settings.
(iv) C-Flat++ achieves performance comparable to C-Flat but utilizes significantly smaller sharpness minimization, leading to a substantial reduction in training time. 

Figure \ref{fig:cflatpp} provides a comparative analysis of SGD, C-Flat, and C-Flat++. As shown in Figure \ref{subfig:cflatpp_acc}, C-Flat consistently outperforms SGD across all tasks, with particularly significant benefits observed in the later tasks. This improvement is due to the fact that the loss landscape optimized by C-Flat is flatter than that of SGD, making it more robust to parameter changes during updates. 
Figure \ref{subfig:cflatpp_speed} shows that C-Flat++ delivers performance comparable to C-Flat, while being significantly faster. This speedup can be attributed to the adaptive sharpness minimization policy, as illustrated in Figure \ref{subfig:cflatpp_prop}. Furthermore, it is noteworthy that C-Flat++ places greater emphasis on sharpness in the later iterations of each milestone, a behavior consistent with our findings in Table \ref{tab:subcflat} and Subsection \ref{subsec:cflatpp}. 
Although the evaluation of sharpness is less comprehensive, it does not impede the minimization of sharpness in the loss landscape, as demonstrated in Figure \ref{subfig:cflatpp_snorm}.

In summary, C-Flat and C-Flat++ strengthen baseline methods across all CL categories, serving as valuable enhancements that complement and extend existing approaches.

% \begin{figure}[t]
% \centering
%     \subfloat[MEMO Epoch: 50]{\includegraphics[width=1.5in]{main_figs/esd_base_50.pdf}%
%         \label{fig_esd_50_wo}}
%     \hfil
%     \subfloat[MEMO w/ C-Flat]{\includegraphics[width=1.5in]{main_figs/esd_cflat_50.pdf}%
%         \label{fig_esd_50_w}}
%     \hfil
%     \subfloat[MEMO Epoch: 150]{\includegraphics[width=1.5in]{main_figs/esd_base_150.pdf}%
%         \label{fig_esd_150_wo}}
%     \hfil
%     \subfloat[MEMO w/ C-Flat ]{\includegraphics[width=1.5in]{main_figs/esd_cflat_150.pdf}%
%         \label{fig_esd_150_w}}
%     \hfil
% \caption{The Hessian eigenvalues and the traces at epochs 50, and 150 on B0\_Inc10 setting (MEMO, CIFAR-100) w/ and w/o C-Flat plugged in.}
% \label{fig_esd}
% \end{figure}

\begin{figure}[t]
\centering
    \subfloat[MEMO Epoch: 50]{\includegraphics[width=1.7in]{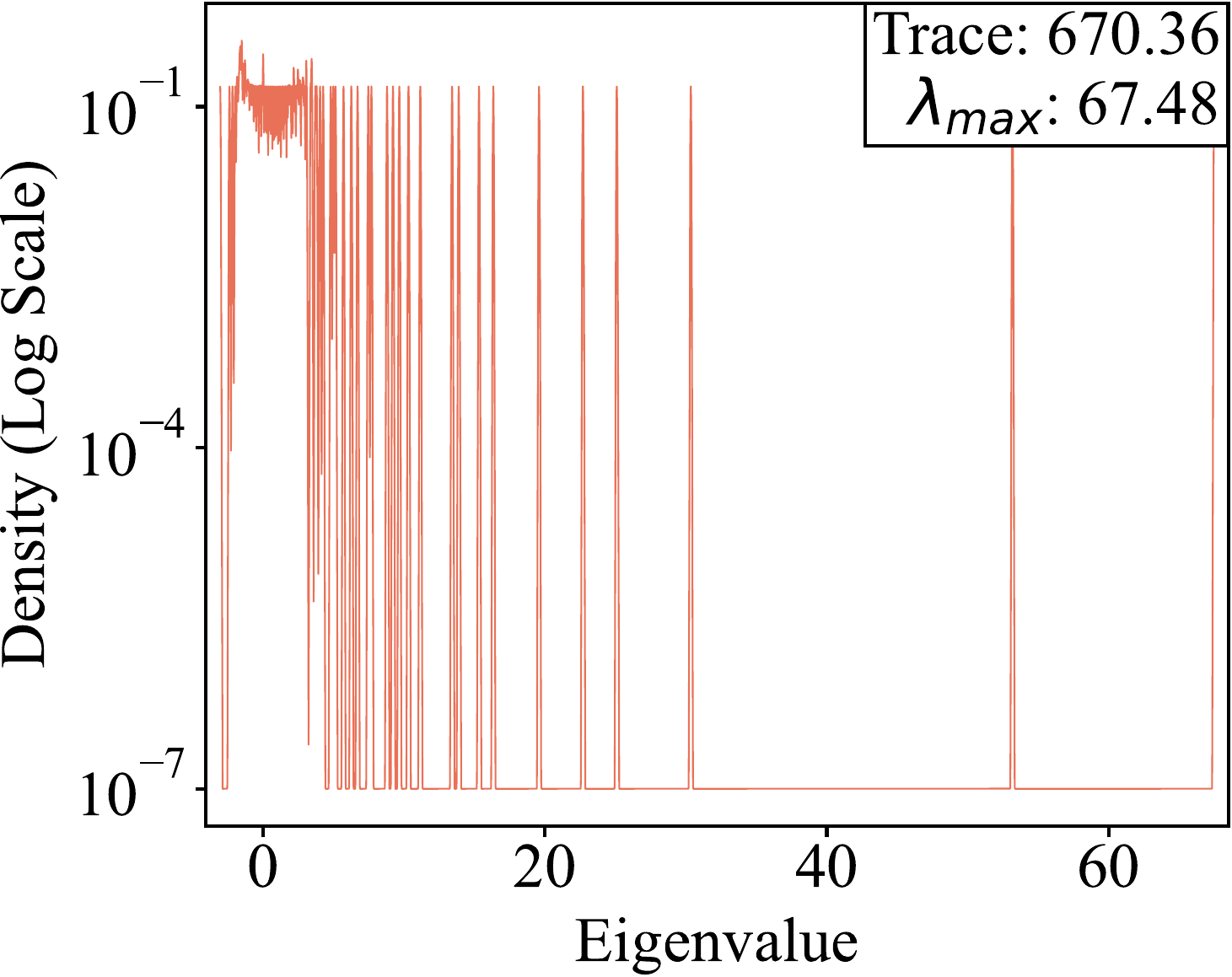}%
        \label{fig_esd_50_base}}
    \hfil
    \subfloat[MEMO Epoch: 150]{\includegraphics[width=1.7in]{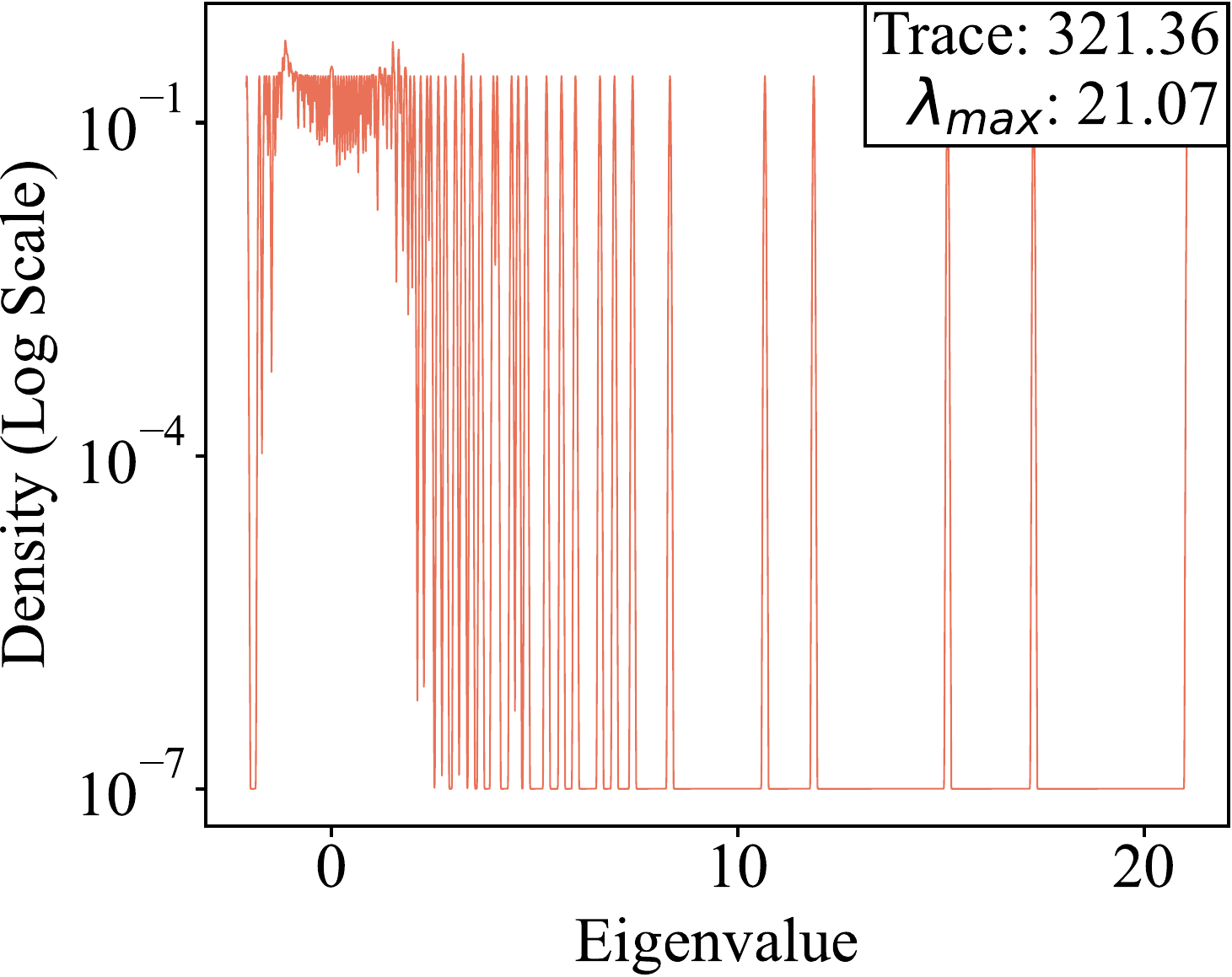}%
        \label{fig_esd_150_base}}
    \hfil
    \subfloat[MEMO w/ C-Flat]{\includegraphics[width=1.7in]{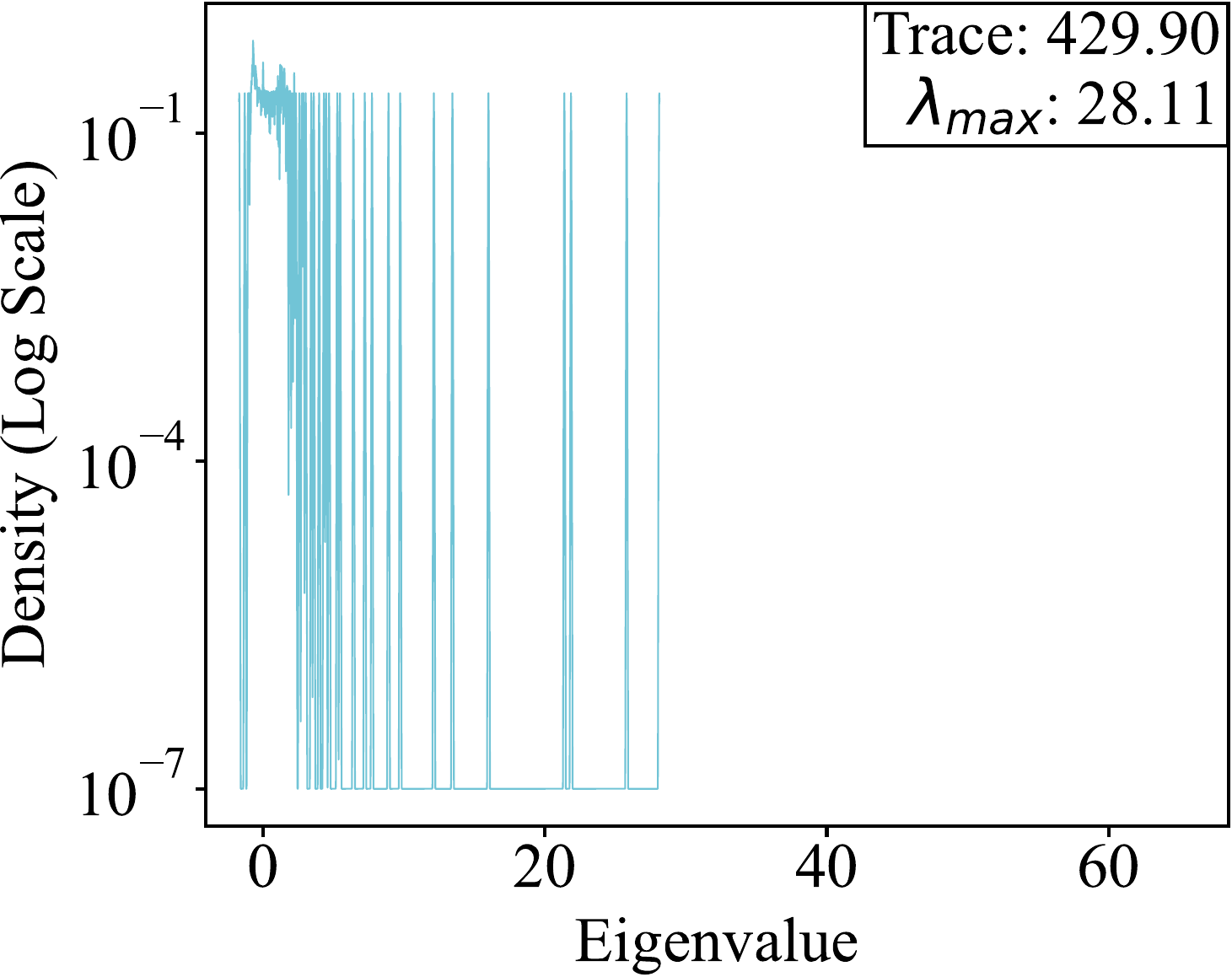}%
        \label{fig_esd_50_cflat}}
    \hfil
    \subfloat[MEMO w/ C-Flat]{\includegraphics[width=1.7in]{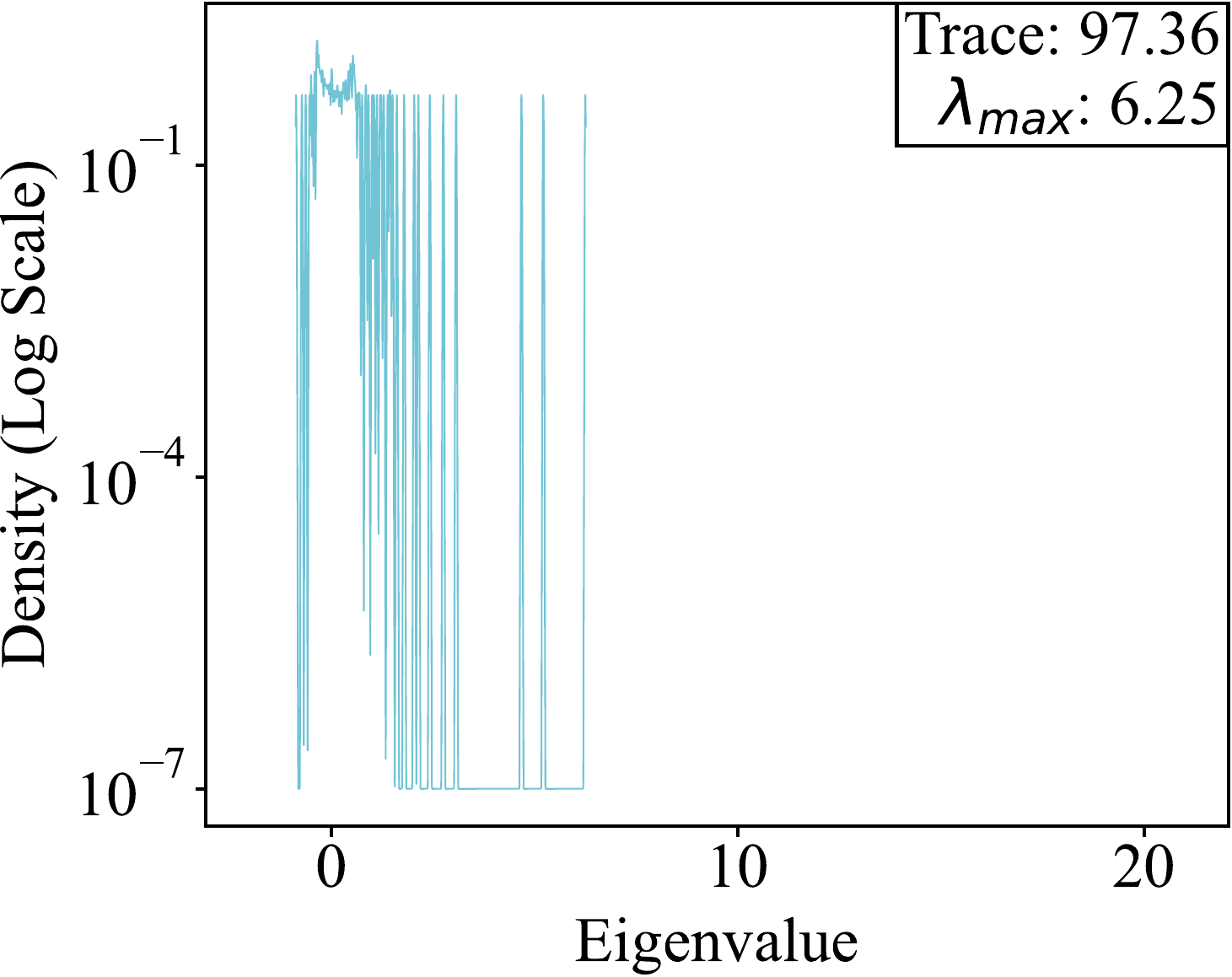}%
        \label{fig_esd_150_cflat}}
    \hfil
    \subfloat[MEMO w/ C-Flat++]{\includegraphics[width=1.7in]{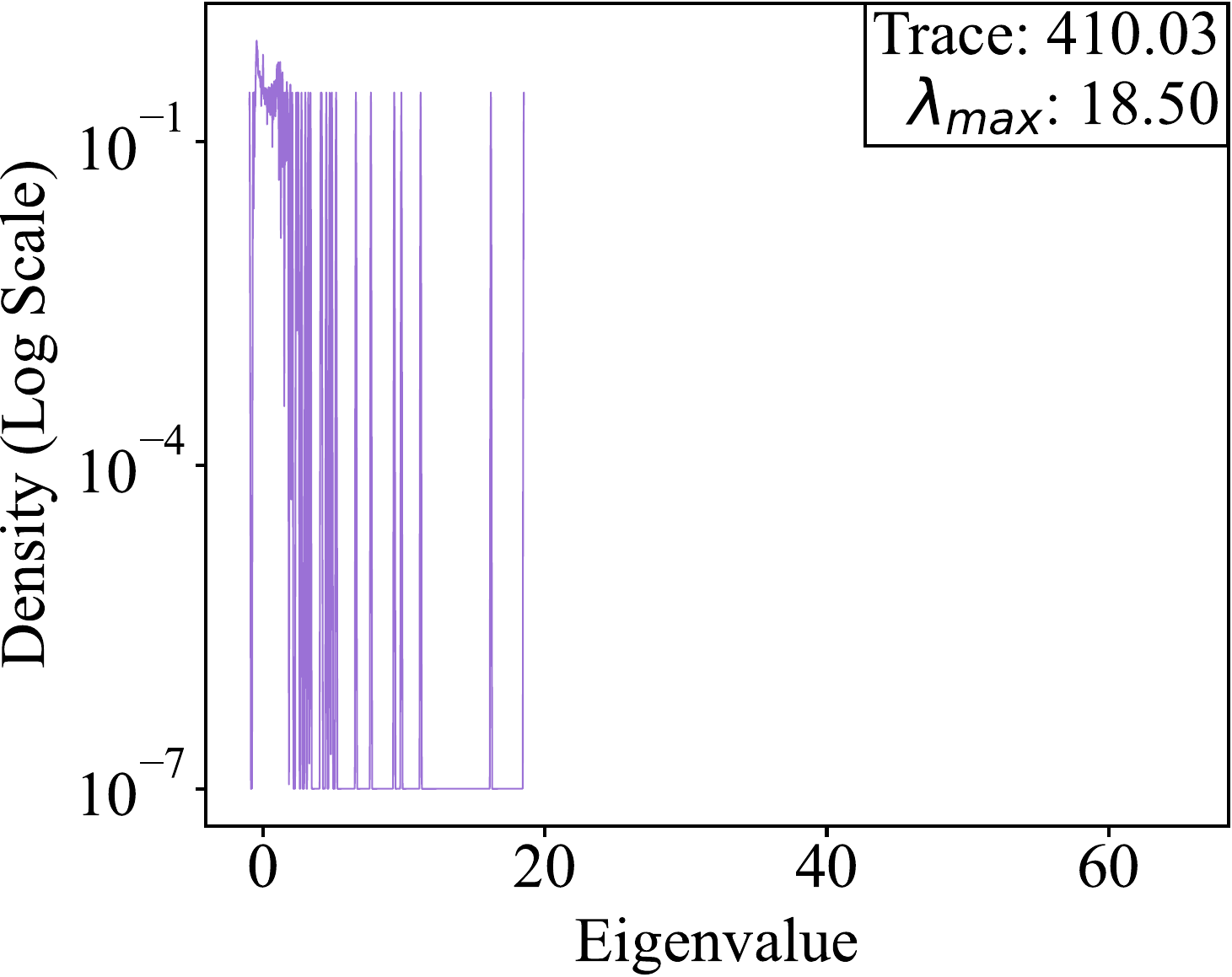}%
        \label{fig_esd_50_cflat++}}
    \hfil
    \subfloat[MEMO w/ C-Flat++]{\includegraphics[width=1.7in]{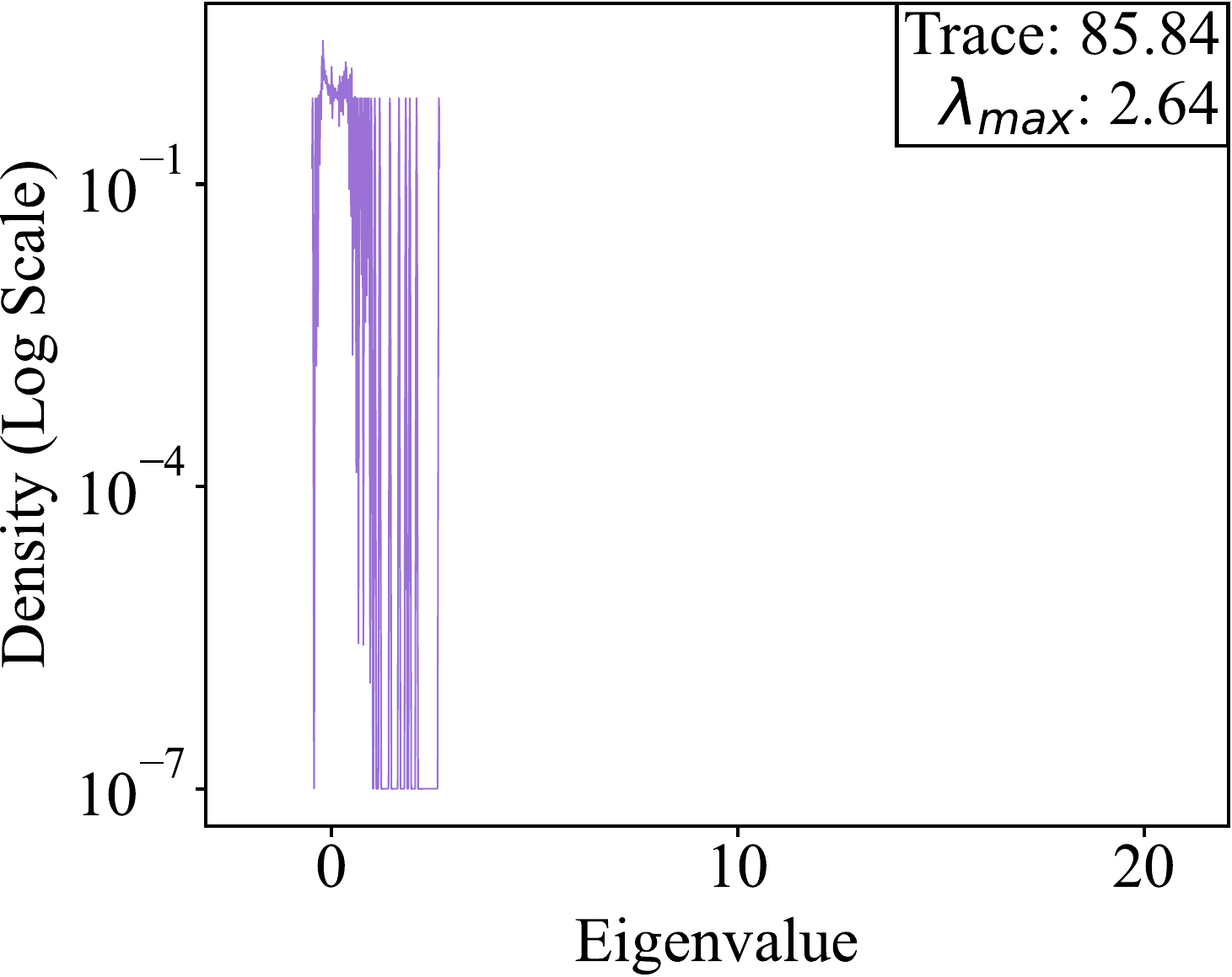}%
        \label{fig_esd_150_cflat++}}
    \hfil
\caption{The Hessian eigenvalues and the traces at epochs 50, and 150 on B0\_Inc10 setting (MEMO, CIFAR-100) w/ and w/o C-Flat / C-Flat++ plugged in.}
\label{fig_esd}
\end{figure}

\subsection{Hessian Eigenvalues and Hessian Traces}

\textbf{Hessian Eigenvalues.} Equation~\ref{eq:Hessian} establishes the link between first-order flatness and Hessian eigenvalues in CL. Generally, Hessian eigenvalues serve as a key metric for assessing the sharpness of the loss landscape. To empirically analyze this, we report Hessian eigenvalue distributions in Figure~\ref{fig_esd}. As shown, models trained with vanilla-SGD exhibit significantly higher maximal Hessian eigenvalues (67.48/21.07 at epochs 50/150 in Figure~\ref{fig_esd_50_base} and Figure~\ref{fig_esd_150_base}), indicating a failure to reduce curvature effectively, which is suboptimal for CL. In contrast, C-Flat induces a substantial reduction in Hessian eigenvalues to 28.11/6.25 at epochs 50/150 (Figure~\ref{fig_esd_50_cflat} and Figure~\ref{fig_esd_150_cflat}), leading to flatter minima and improved CL performance. Moreover, Figure~\ref{fig_esd_50_cflat++} and Figure~\ref{fig_esd_150_cflat++} reveal that C-Flat++, leveraging an efficient proxy, further reduces Hessian eigenvalues to 18.50/2.64 at epochs 50/150, reinforcing its effectiveness in enhancing flatness.

\textbf{Hessian Traces.} To quantify the sharpness of the loss landscape at convergence, we approximate the Hessian using the empirical Fisher information matrix and analyze its trace. As depicted in Figure~\ref{fig_esd}, our method achieves a substantial reduction in Hessian trace compared to vanilla-SGD (670.36/321.36 drops to 429.90/97.36 and 410.03/85.84 at epochs 50/150 in Figure~\ref{fig_esd_50_cflat} to Figure~\ref{fig_esd_150_cflat++}), indicating a flatter minimum. These empirical findings strongly align with and substantiate the theoretical insights discussed in the methodology section.

\subsection{Visualization of Landscapes}
\label{landscapes}

% \setlength{\intextsep}{0pt}
% \begin{wrapfigure}{l}{0.48\textwidth}
%   \centering
%    \subfloat[B0\_Inc10]{\includegraphics[width=1.3in]{main_figs/sam_clat_10.pdf}}
%     \hfil
%     \subfloat[B0\_Inc20]{\includegraphics[width=1.3in]{main_figs/sam_clat_20.pdf}}
%     \hfil
% \caption{C-Flat vs. Zero-order flatness}
% \label{fig_zero}
% \end{wrapfigure}

% \begin{wrapfigure}{l}{0.48\textwidth}
%   \centering
%   \subfloat[Convergence speed
%   % on Replay with CIFAR-100/B0\_Inc20.
%   \label{subfig:converge_seed}]{\includegraphics[width=1.4in]{main_figs/converge_speed.pdf}}
%   % \quad
%   \subfloat[Runtime 
%   % (From left to right, the ratios are 20\%/50\%/100\%.)
%   \label{subfig:runtime_acc}]{\includegraphics[width=1.4in]{main_figs/runtime_acc.pdf}}
%   \caption{Analysis of computation overhead}
%   \label{fig:runtime}
%   % \vspace{1em}
% \end{wrapfigure}

More intuitively, we provide a detailed visualization of the loss landscape. PyHessian~\cite{DBLP:journals/corr/abs-1912-07145} is used to depict the loss landscape of models. For simplicity, we select one representative method from each category of CL approaches (Replay, WA, MEMO) for evaluation. Figure~\ref{fig_loss_landscape} clearly demonstrates that applying C-Flat results in a significantly flatter loss landscape compared to the vanilla method. This trend consistently holds across different CL categories, providing strong empirical support for C-Flat and reinforcing our intuition.
Beyond this, the loss landscape of C-Flat++ exhibits a similarly flat region compared to the baseline models. Moreover, in certain critical eigenvector directions, C-Flat++ achieves even lower losses, owing to its soft regularization constraint, which is implemented via a mixed optimizer combining C-Flat and vanilla optimization.

% \begin{wrapfigure}{l}{0.48\textwidth}
%   \centering
%    \subfloat[B0\_Inc10]{\includegraphics[width=1.3in]{main_figs/sam_clat_10.pdf}}
%     \hfil
%     \subfloat[B0\_Inc20]{\includegraphics[width=1.3in]{main_figs/sam_clat_20.pdf}}
%     \hfil
% \caption{C-Flat vs. Zero-order flatness}
% \label{fig_zero}
% \end{wrapfigure}

\begin{figure}[t]
\centering
    \subfloat[B0\_Inc10]{\includegraphics[width=1.7in]{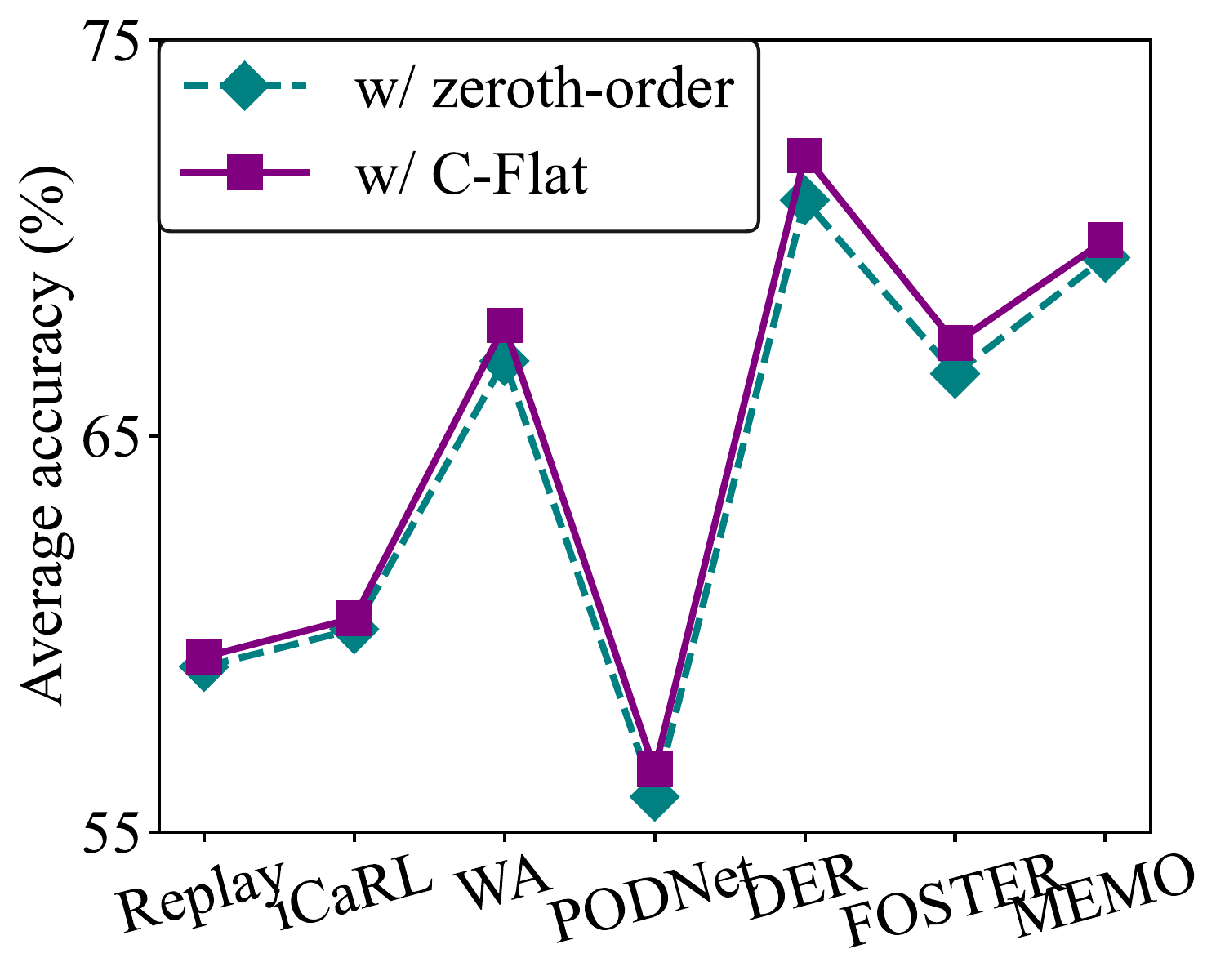}}
    \hfil
    \subfloat[B0\_Inc20]{\includegraphics[width=1.7in]{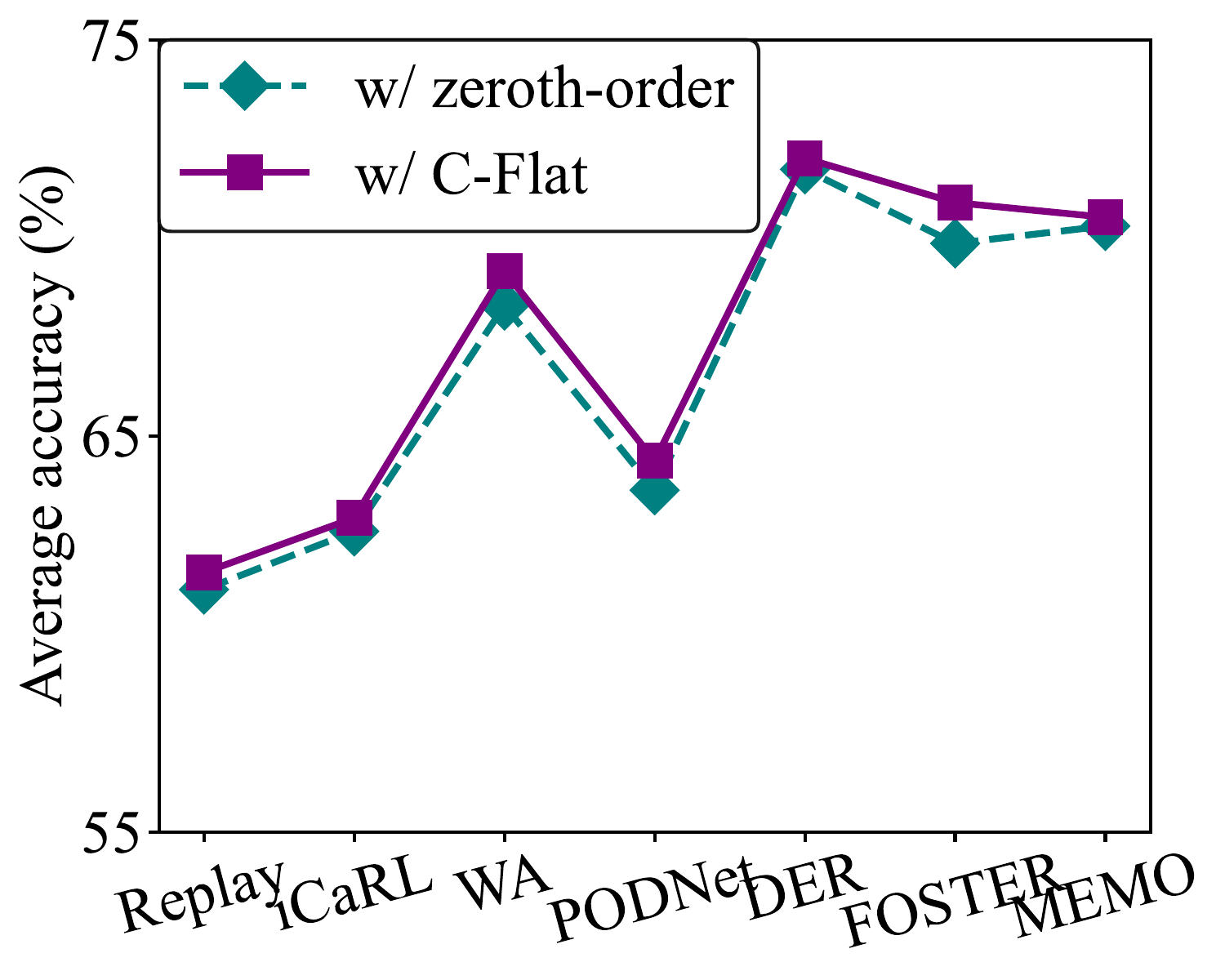}}
\caption{C-Flat vs. Zero-order flatness.}
\label{fig_zero}
\end{figure}

\begin{table}[t]
\scriptsize
\centering
\caption{Revisiting FS-DGPM series using C-Flat and C-Flat++. Bold indicates the best results and underline denotes the second-best.}
\label{tbl_fs_transposed_no_w_cflat}
\begin{tblr}{
  colspec = {l ccccc},
  % cell{5}{1} = {r=2}{},
  hline{1,2,5} = {-}{},
}
Method     & La-GPM    & FS-GPM    & DGPM      & La-DGPM   & FS-DGPM   \\
% \hline
Oracle~\cite{deng2021flattening}     & 72.90     & 73.12     & 72.66     & 72.85     & 73.14     \\
w/ C-Flat~\cite{bian2025make}  & \textbf{73.66}     & \textbf{73.57}     & \textbf{73.01}     & \textbf{73.64}     & \textbf{73.72}     \\
% \hline
w/ C-Flat++  & \underline{73.49}     & \underline{73.40}     & \underline{72.92}     & \underline{73.53}     & \underline{73.66}    \\
% \textit{Boost}       & \textcolor{red}{\textbf{+0.76}}     & \textcolor{red}{\textbf{+0.45}}     & \textcolor{red}{\textbf{+0.35}}     & \textcolor{red}{\textbf{+0.79}}     & \textcolor{red}{\textbf{+0.58}}   \\
%                      & \textcolor{red}{\textbf{+0.76}}     & \textcolor{red}{\textbf{+0.45}}     & \textcolor{red}{\textbf{+0.35}}     & \textcolor{red}{\textbf{+0.79}}     & \textcolor{red}{\textbf{+0.58}}   
\end{tblr}
\end{table}

\subsection{Revisiting Zeroth-order Sharpness}

Limited work~\cite{deng2021flattening, shi2021overcoming} proved that the zeroth-order sharpness leads to flat minima boosted CL. Here, we employ a zeroth-order optimizer~\cite{foret2020sharpness} instead of vanilla-SGD to verify the performance of C-Flat. As shown in Figure~\ref{fig_zero}, C-Flat (purple line) stably outperforms the zeroth-order sharpness (blue line) on all baselines. We empirically demonstrated that flatter is better for CL.

Former work FS-DGPM~\cite{deng2021flattening} regulates the gradient direction with flat minima to promote CL. The FS (Flattening Sharpness) term derived from FS-DGPM is a typical zeroth-order flatness. We revisit the FS-DGPM series (including La/FS-GPM, DGPM, La/FS-DGPM)~\cite{deng2021flattening, saha2020gradient} to evaluate performance using C-Flat instead of FS (see Algorithm~\ref{alg:FS}). Table~\ref{tbl_fs_transposed_no_w_cflat} yields three conclusions: (i) C-Flat boosts the GPM~\cite{saha2020gradient} baseline as a pluggable regularization term. This not only extends the frontiers of CL methods, incorporating gradient-based solutions, but also reaffirms the remarkable versatility of C-Flat. (ii) Throughout all series of FS-DGPM, C-Flat seamlessly supersedes FS and achieves significantly better performance. This indicates that C-Flat consistently exceeds zeroth-order sharpness. (iii) C-Flat++ with partial sharpness minimization in later iterations is sufficient for enhancing GPM. Hence, reconfirming that C-Flat++ is indeed a simple yet potential CL method that deserves to be widely spread within the CL community.

\begin{figure*}
\vspace{-4mm}
\centering
    \subfloat[$\lambda$ on CL methods\label{subfig:param_lambda_CL}]{\includegraphics[width=1.7in]{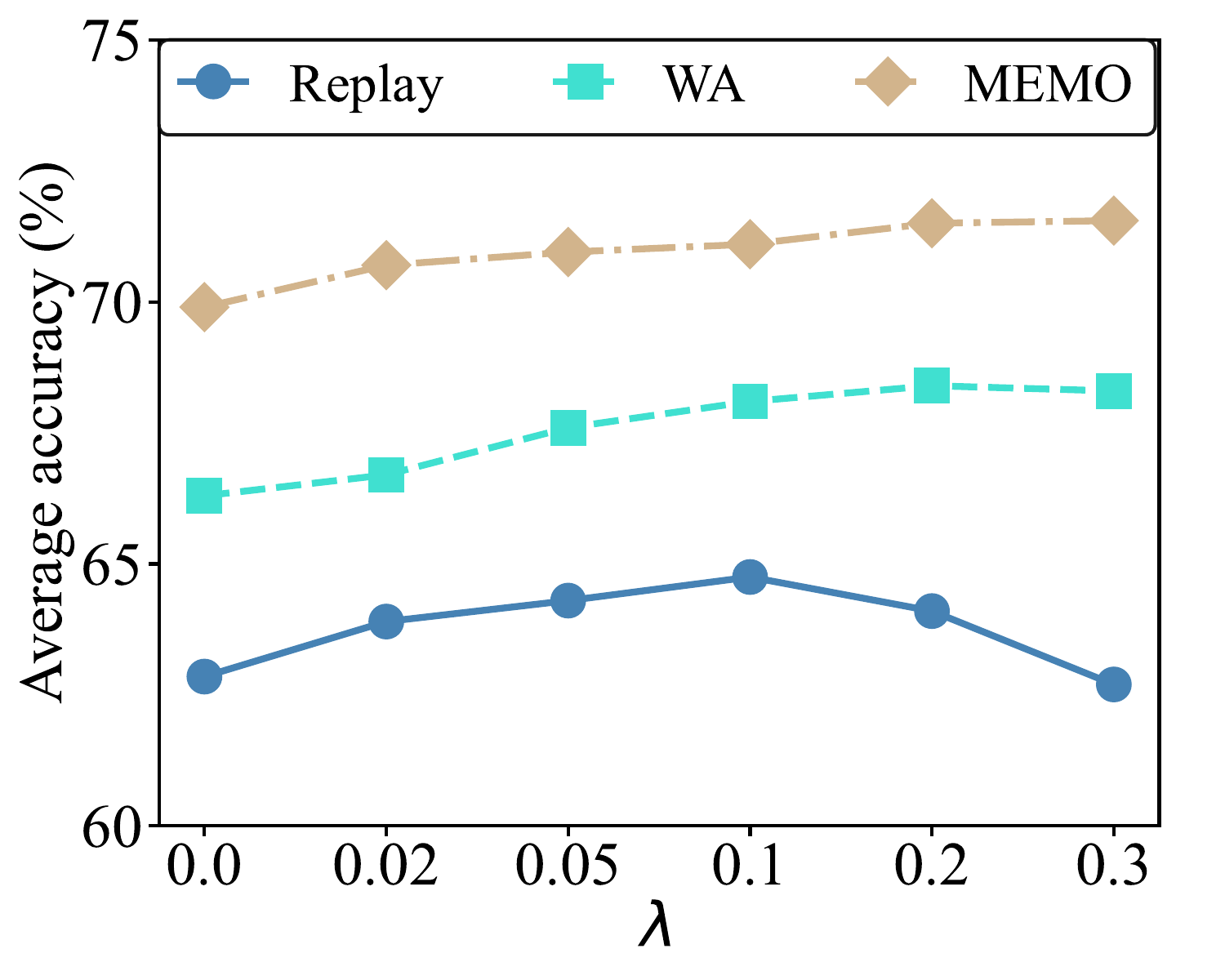}}
    % \hfil
    \subfloat[$\rho$ on CL methods\label{subfig:param_rho_CL}]{\includegraphics[width=1.7in]{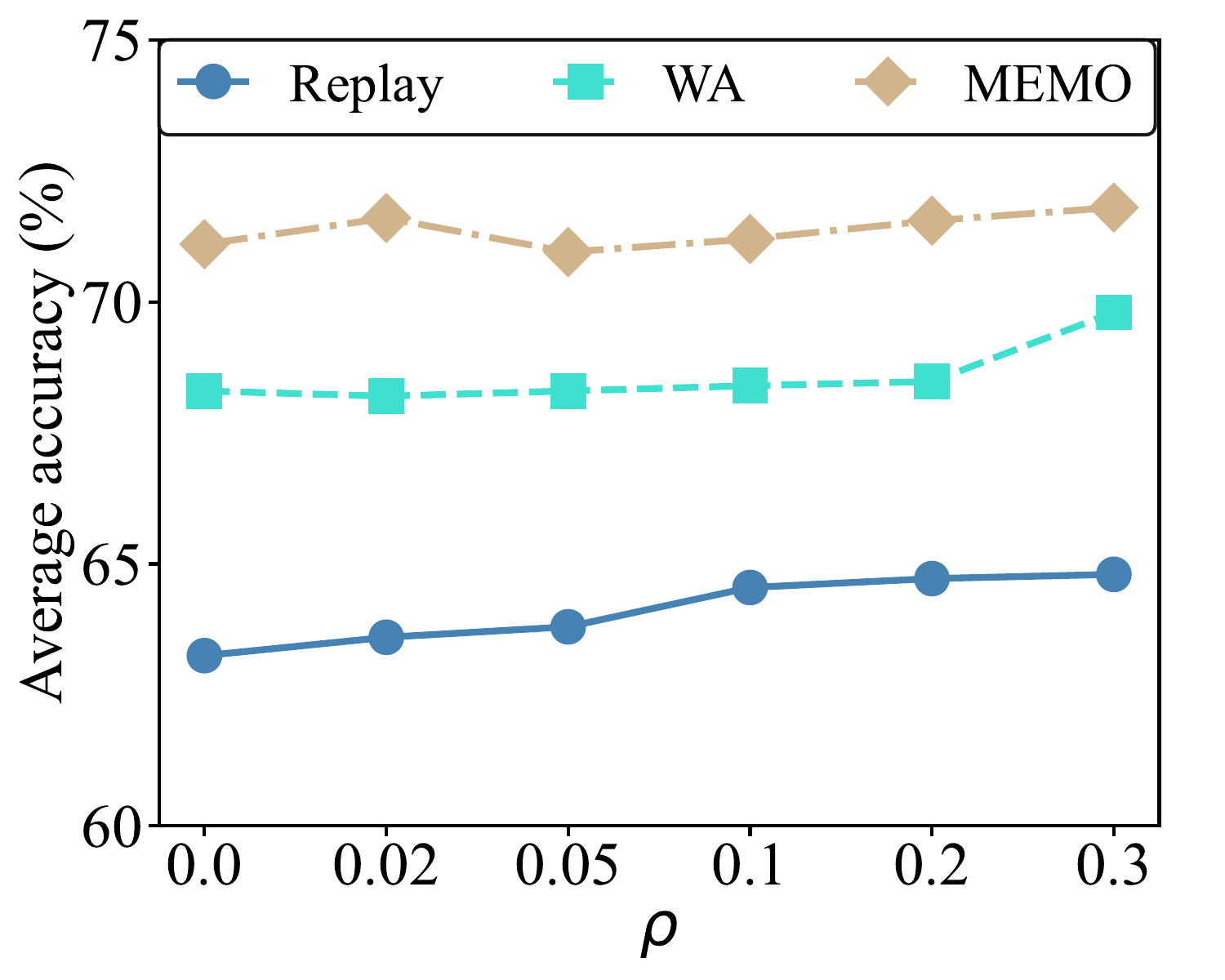}}
    % \hfil
    \subfloat[
    Effects of $\rho$
    % $\rho$ on flat sharpness optimizers
    \label{subfig:param_rho_optim}]{\includegraphics[width=1.7in]{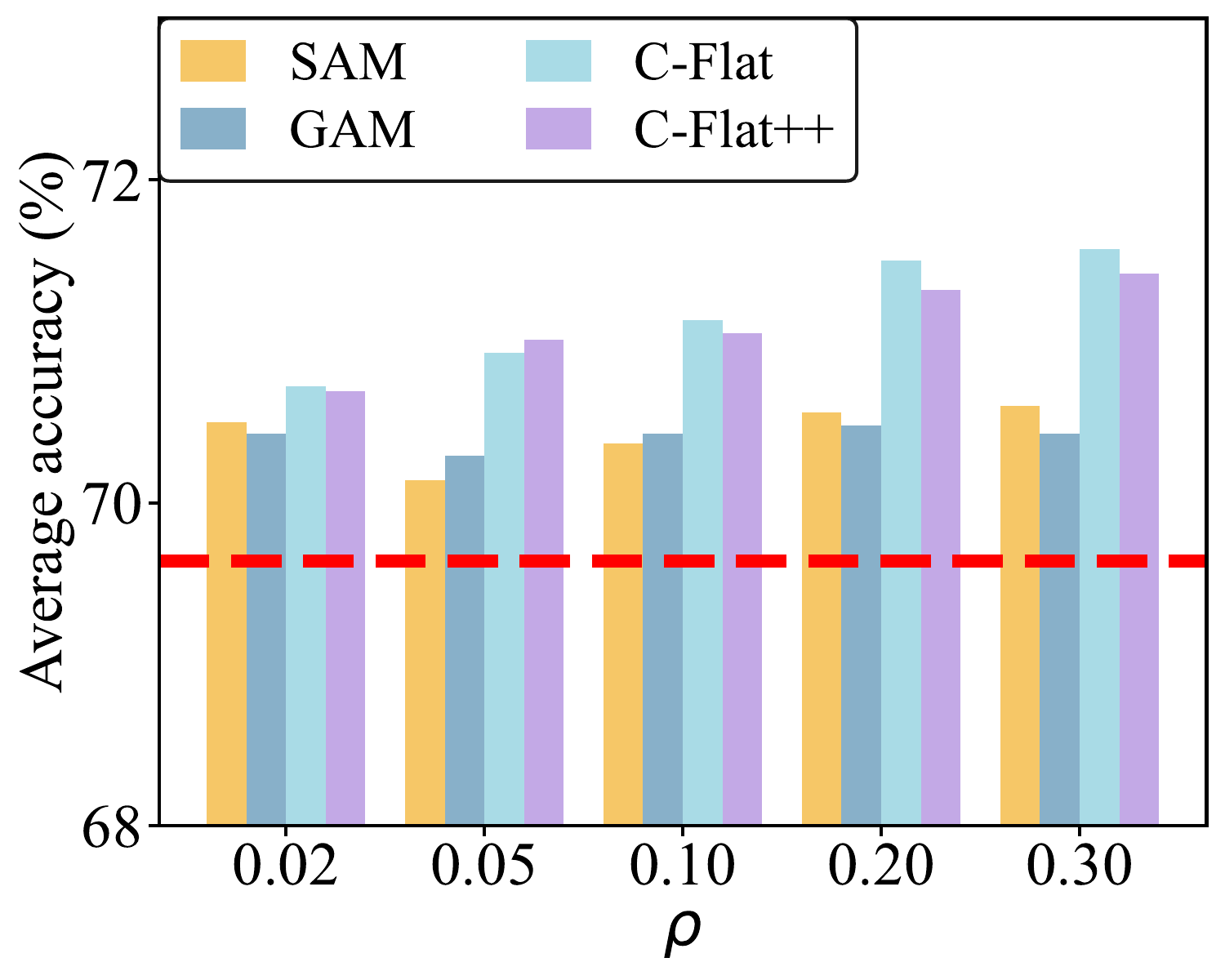}}
    % \hfil
    \subfloat[
    Effects of $\rho$ scheduler
    % $\rho$ scheduler on flat sharpness optimizers
    \label{subfig:param_rho_sche}]{\includegraphics[width=1.7in]{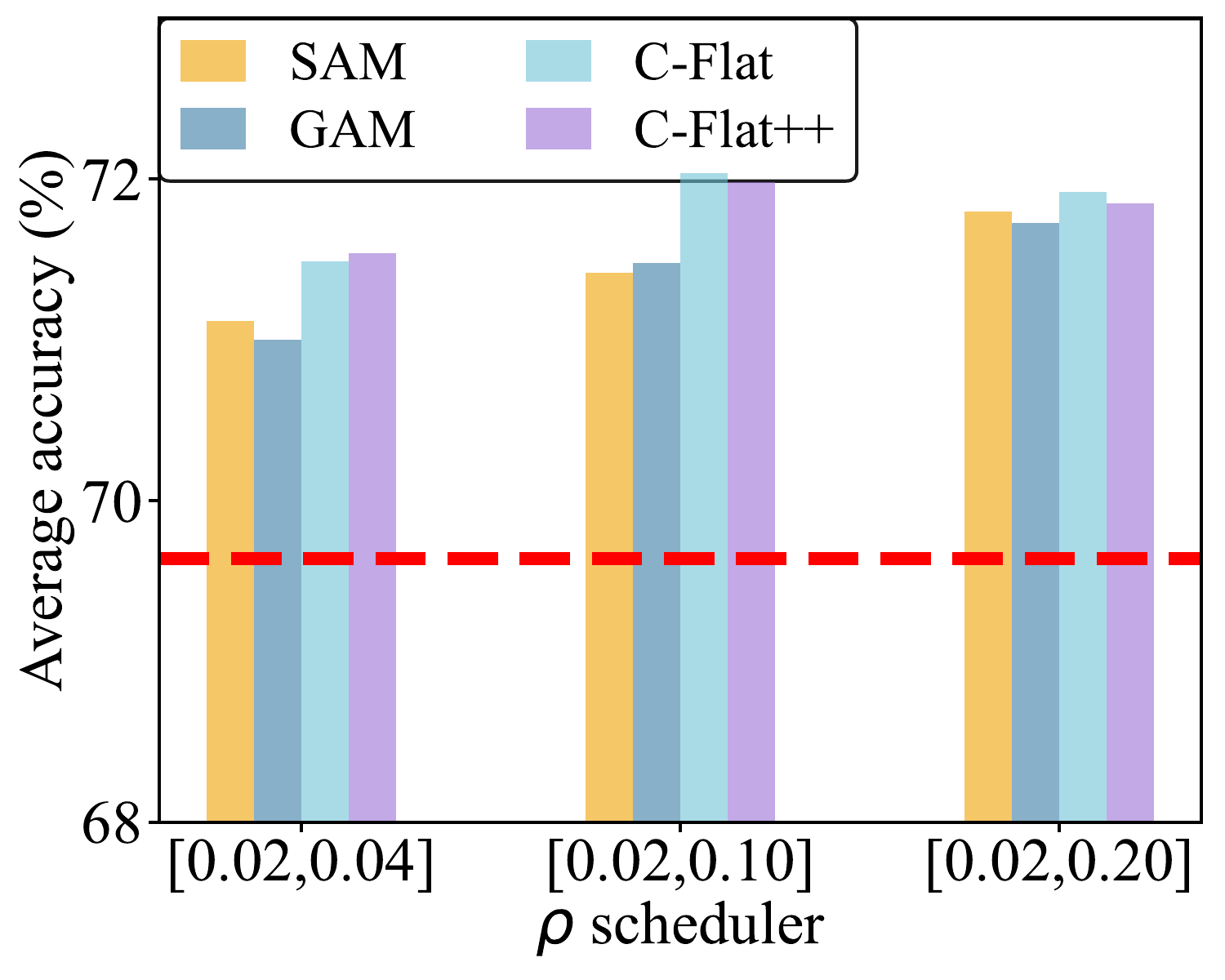}}
    \hfil
    \caption{Ablation studies. (a) and (b) illustrate the effect of $\lambda$ and $\rho$ on different CL methods (WA, Replay, MEMO). (c) and (d) show the impact of $\rho$ and the $\rho$ scheduler on MEMO with different optimizers (SGD (red line), SAM, GAM, C-Flat, C-Flat++).}  
  \vspace{-4mm}
\label{fig:params}
\end{figure*}

\begin{figure}
\vspace{-4mm}
\centering
    \subfloat[Convergence analysis]{\includegraphics[width=1.7in]{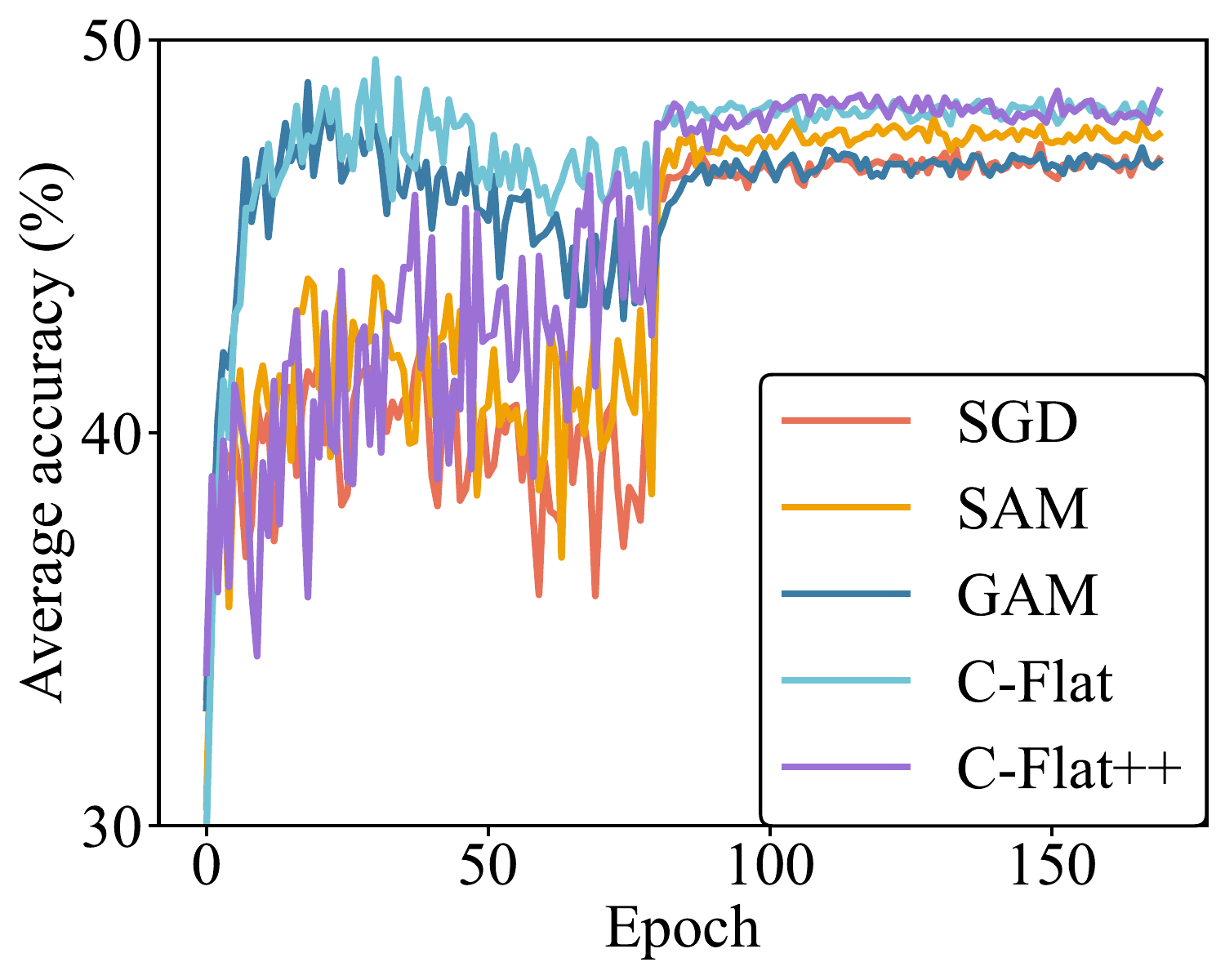}%
        \label{subfig:converge_seed}}
    \hfil
    \subfloat[Training time]{\includegraphics[width=1.7in]{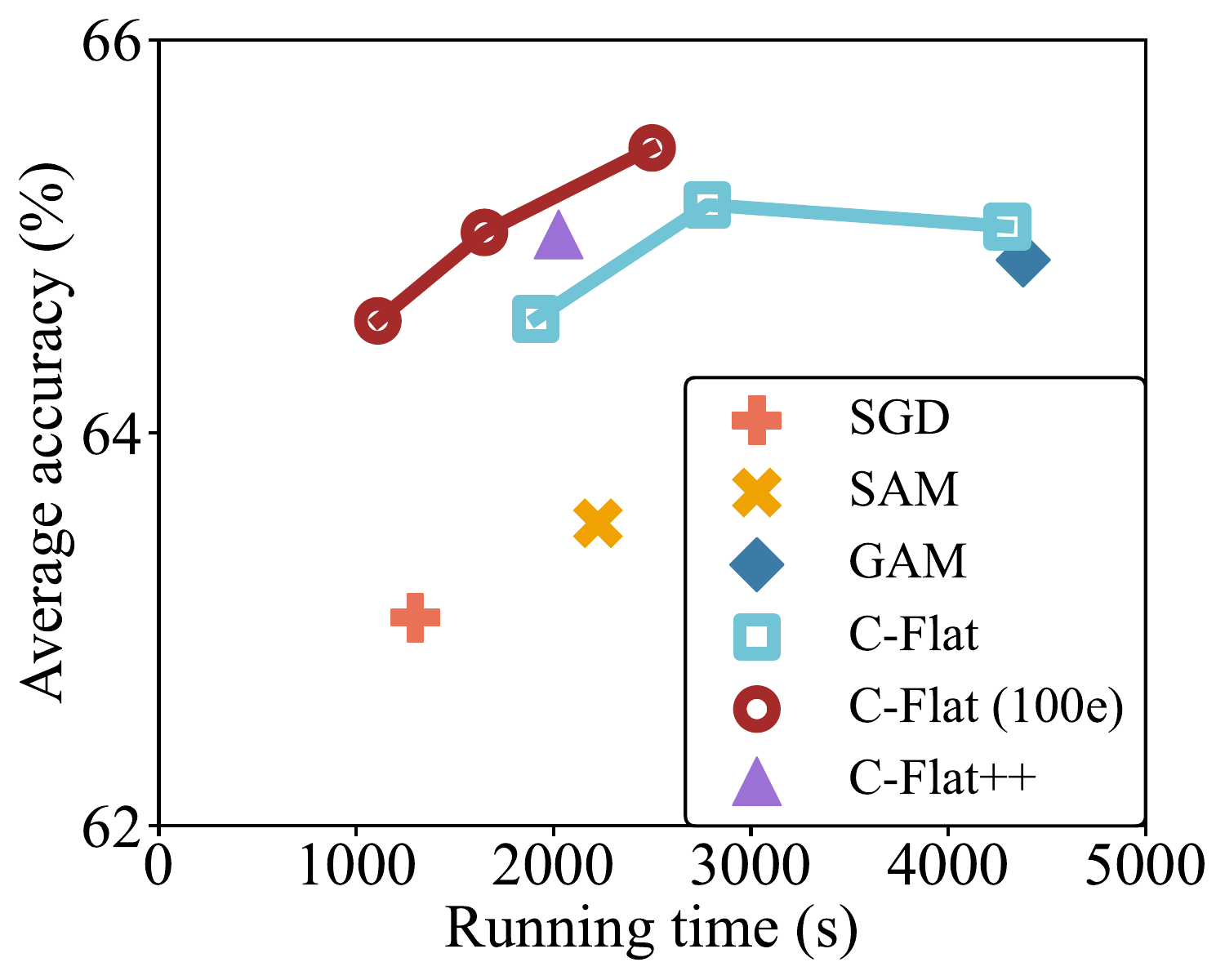}%
        \label{subfig:runtime_acc}}
\caption{Analysis of computation overhead.}
\label{fig:runtime}
\end{figure}

\begin{table}[t]
\centering
\caption{A tiered guideline of C-Flat. Boost indicates the enhancement brought by C-Flat over others.}
\label{tbl_tier}
\begin{tabular}{@{}ccc@{}}
\toprule
Level & Speed                                    & Boost (SGD/SAM/C-Flat++)  \\ \midrule
L1    & SGD \textgreater{} C-Flat++ \textgreater{} SAM \textgreater{} \textbf{C-Flat} & \textcolor{red}{+2.39\%}/\textcolor{red}{+1.91\%}/\textcolor{red}{+0.34\%} \\
L2    & SGD \textgreater{} \textbf{C-Flat} \textgreater{} C-Flat++ \textgreater{} SAM & 
\textcolor{red}{+1.52\%}/\textcolor{red}{+1.04\%}/\textcolor{forestgreen}{-0.53\%} \\
L3    & \textbf{C-Flat} \textgreater{} SGD \textgreater{} C-Flat++ \textgreater{} SAM & \textcolor{red}{+1.51\%}/\textcolor{red}{+1.03\%}/\textcolor{forestgreen}{-0.54\%} \\ \bottomrule
\end{tabular}
\end{table}

\subsection{Computation Overhead}

This subsection provides a thorough analysis of the computational overhead of C-Flat and C-Flat++ in comparison to other sharpness-aware optimizers. Notably, in our conference paper, we train C-Flat on CL benchmarks using 20\%, 50\%, and 100\% of the total iterations (\textit{every 5, 2, and 1 steps}), and approximately 60\% of the epochs for efficiency, as it converges faster than other optimizers. 
Figure \ref{subfig:converge_seed} demonstrates that C-Flat converges the fastest and achieves the highest accuracy (purple line), indicating that only a few iterations or epochs with C-Flat are necessary to improve CL performance. Later, C-Flat++ also converges to similar performance as C-Flat.
Figure~\ref{subfig:runtime_acc} shows the following:
i) Compared to SGD, C-Flat with only 20\% of the iterations and 60\% of the epochs (brown line) achieves better CL performance in slightly less time;
ii) C-Flat outperforms GAM, achieving similar results to SAM when the iterations/epochs ratio is set to 50\%/60\%;
iii) Models trained with C-Flat for 100 epochs outperform those trained with other optimizers for 170 epochs;
iv) C-Flat++ using approximately 24\% of C-Flat's optimization significantly surpasses other sharpness-aware optimizers in terms of both running speed and accuracy.

To discuss practicality better, we provided a tier guideline, which categorizes C-Flat into L1 to L3 levels, as shown in Table~\ref{tbl_tier}, L1 denotes the low-speed version of C-Flat, with a slightly lower speed than SAM and the best performance; L2 follows next; L3 denotes the high-speed version of C-Flat, with a faster speed than SGD and a performance close to L2. Beyond this, C-Flat++ serves as a practical solution for achieving both high accuracy and efficiency.

\subsection{Ablation Study}

We perform ablation study in two cases: (i) the influence of $\lambda$ and $\rho$ on different CL methods; (ii) the influence of $\rho$ and its scheduler on different optimizers. 
% We analyze the influence from both hyperparameters in the following aspects.

% \textbf{hyperparameters $\lambda$.} To scrutinize the effect of $\lambda$, we first present the performance of C-Flat with varying $\lambda$. As described in equation 13 of the main paper, $\lambda$ controls the strength of the C-Flat penalty. 
% When $\lambda$ is equal to 0, this means that first-order flatness is not implemented. 
% As shown in Figure~\ref{fig:params} (a), (b) and (c), compared with vanilla optimizer, C-Flat shows remarkable improvement with varying $\lambda$. 

% \textbf{hyperparameters $\rho$.} To scrutinize the effect of $\rho$, we present the performance of C-Flat with varying $\rho$. $\rho$ controls controls the step length of gradient ascent. We plot the performance of different CL method with varying $\rho$. As shown in Figure~\ref{fig:params} (d), (e) and (f), C-Flat with $\rho$ larger than 0 outperforms C-Flat without gradient ascent, showing that the gradient ascent is necessary for C-Flat.

% \textbf{hyperparameters $\lambda$ and $\rho$ on CL methods.} 
% \textbf{How $\lambda$ and $\rho$ affects various CL methods?}
We first present the performance of C-Flat with varying $\lambda$ and $\rho$. As described in Eq. 13, $\lambda$ controls the strength of the C-Flat penalty (when $\lambda$ is equal to 0, this means that first-order flatness is not implemented). 
As shown in Figure~\ref{subfig:param_lambda_CL}, compared with vanilla optimizer, C-Flat shows remarkable improvement with varying $\lambda$. Moreover, $\rho$ controls the step length of gradient ascent. As shown in Figure~\ref{subfig:param_rho_CL}, C-Flat with $\rho$ larger than 0 outperforms C-Flat without gradient ascent, showing that C-Flat benefits from the gradient ascent.

% \textbf{hyperparameters $\rho$ and its scheduler on flat sharpness optimizers.} 
% \textbf{$\rho$ and $\rho$ scheduler on flat sharpness optimizers.} 
% For each CL task $T$, same learning rate $\eta^T$ and neighborhood size $\rho^T$ initialization are used, and $\rho^T_i\in [\rho_{\_}, \rho_{+}]$ drops with $\eta^T_i$ in epochs by $\rho^T_i = \rho_{\_}+\alpha\eta^T_i$, where $\alpha$ is a constant decided by $\rho_{\_}, \rho_{+}$.
% Fig.~\ref{rho},~\ref{rho_scheduler} presents a comparison on $\rho$ initialization and $\{\rho_{\_}, \rho_{+}\}$ scheduler. C-Flat outperforms across various settings, and is not oversensitive to hyperparameters in a reasonable range. 
For each CL task $T$, same learning rate $\eta^T$ and neighborhood size $\rho^T$ initialization are used. By default, $\rho^T_i\in [\rho_{\_}, \rho_{+}]$ is set as a constant, which decays with respect to the learning rate $\eta^T_i\in [\eta_{\_}, \eta_{+}]$ by $\rho^T_i = \rho_{\_}+\frac{(\rho_{+}-\rho_{\_})}{\eta_{+}-\eta_{\_}}(\eta^T_i-\eta_{\_})$.
%, where $\alpha$ is a constant decided by $\rho_{\_}, \rho_{+}$.
Figure~\ref{subfig:param_rho_optim} and Figure~\ref{subfig:param_rho_sche} present a comparison on $\rho$ initialization and $\{\rho_{\_}, \rho_{+}\}$ scheduler. C-Flat outperforms across various settings, and is not oversensitive to hyperparameters in a reasonable range.

\subsection{Knowledge Transfer}

% As is known to all, forward transfer is the desirable condition for CL.

% \textbf{Backward Transfer} a.k.a the ability to transfer knowledge from a current task to improve performance on a previously learned task.

% \textbf{Forward Transfer} a.k.a the ability to transfer knowledge from previous tasks to improve performance and learning efficiency on a related future task.

% \begin{table}
% \scriptsize
% \centering
% \caption{Analysis of BT and FT. RR refers to Relative Return compared to SGD.}
% \label{tbl_beyond}
% \begin{tblr}{
%   colspec = {l ccccc},
%   cell{1,3,5,7}{1} = {r=2}{},
%   cell{1}{3} = {c=3}{},
%   hline{1,3,5,7,9} = {-}{}
% }
% Method      &      & CIFAR-100/ B0\_Inc5     &           &          &        \\
%             &      & SGD                     & C-Flat    & C-Flat++ & RR     \\
% iCaRL~\cite{rebuffi2017icarl}      
%             & old & 53.98 & 56.75 & 55.34 & BT\textcolor{red}{\textbf{+2.10\%/+00.00\%}}     \\
%             & new & 87.07 & 88.15 & 88.40 & FT\textcolor{red}{\textbf{+2.43\%/+00.00\%}}     \\
% PODNet~\cite{DBLP:conf/eccv/DouillardCORV20}     
%             & old & 46.32 & 47.44 & 46.77 & BT\textcolor{red}{\textbf{+2.42\%/+00.00\%}}     \\
%            & new  & 62.65 & 64.75 & 68.74 & FT\textcolor{red}{\textbf{+3.35\%/+00.00\%}}     \\
% FOSTER~\cite{wang2022foster}     
%             & old & 58.50 & 61.35 & 60.28 & BT\textcolor{red}{\textbf{+2.85\%/+00.00\%}}     \\
%             & new & 62.05 & 63.05 & 63.40 & FT\textcolor{red}{\textbf{+1.61\%/+00.00\%}}     \\     
% \end{tblr}
% \end{table}
\begin{figure}[t]
\vspace{-4mm}
\centering
    \subfloat[FWT]{\includegraphics[width=1.7in]{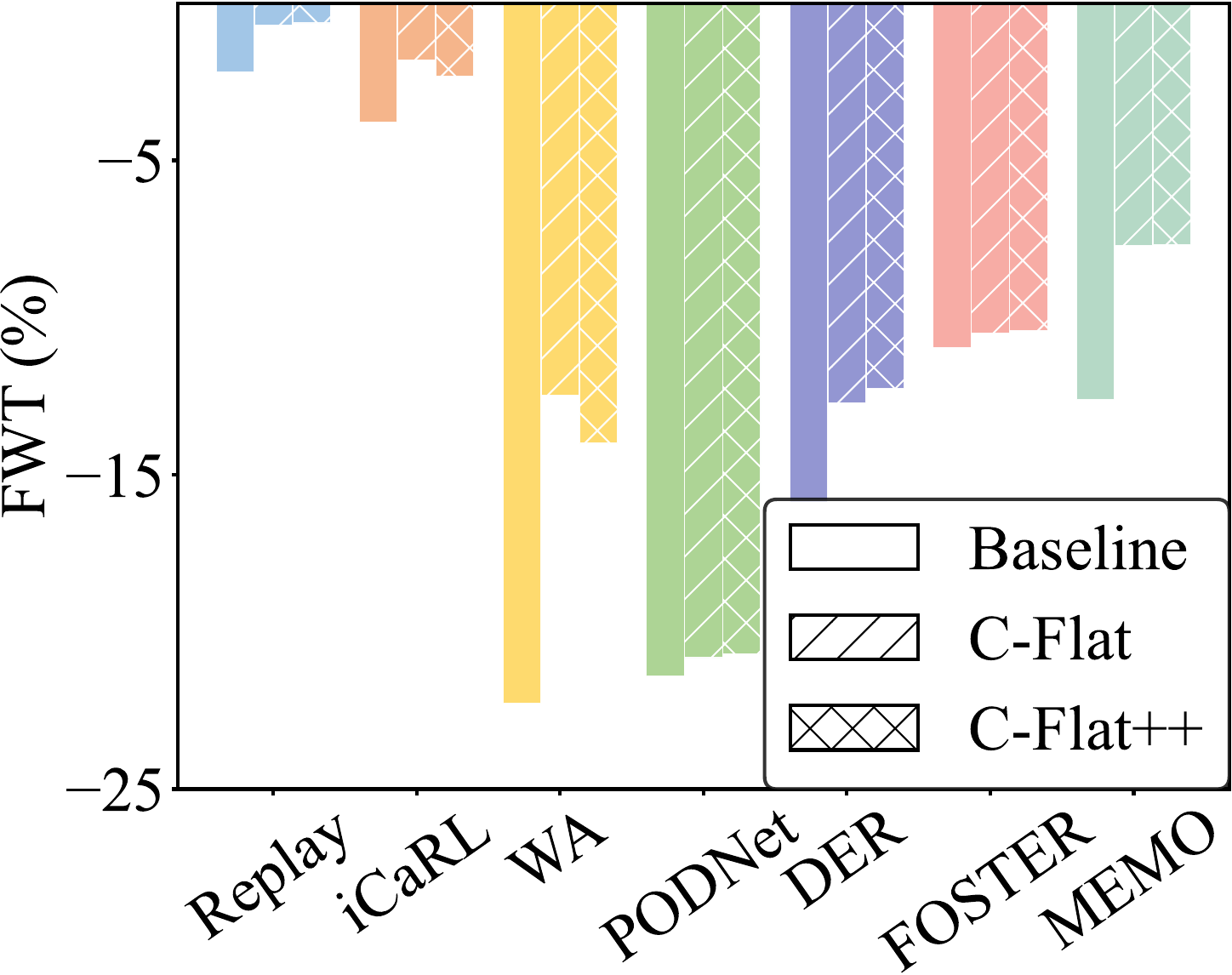}%
        \label{subfig:bwt}}
    \hfil
    \subfloat[BWT]{\includegraphics[width=1.7in]{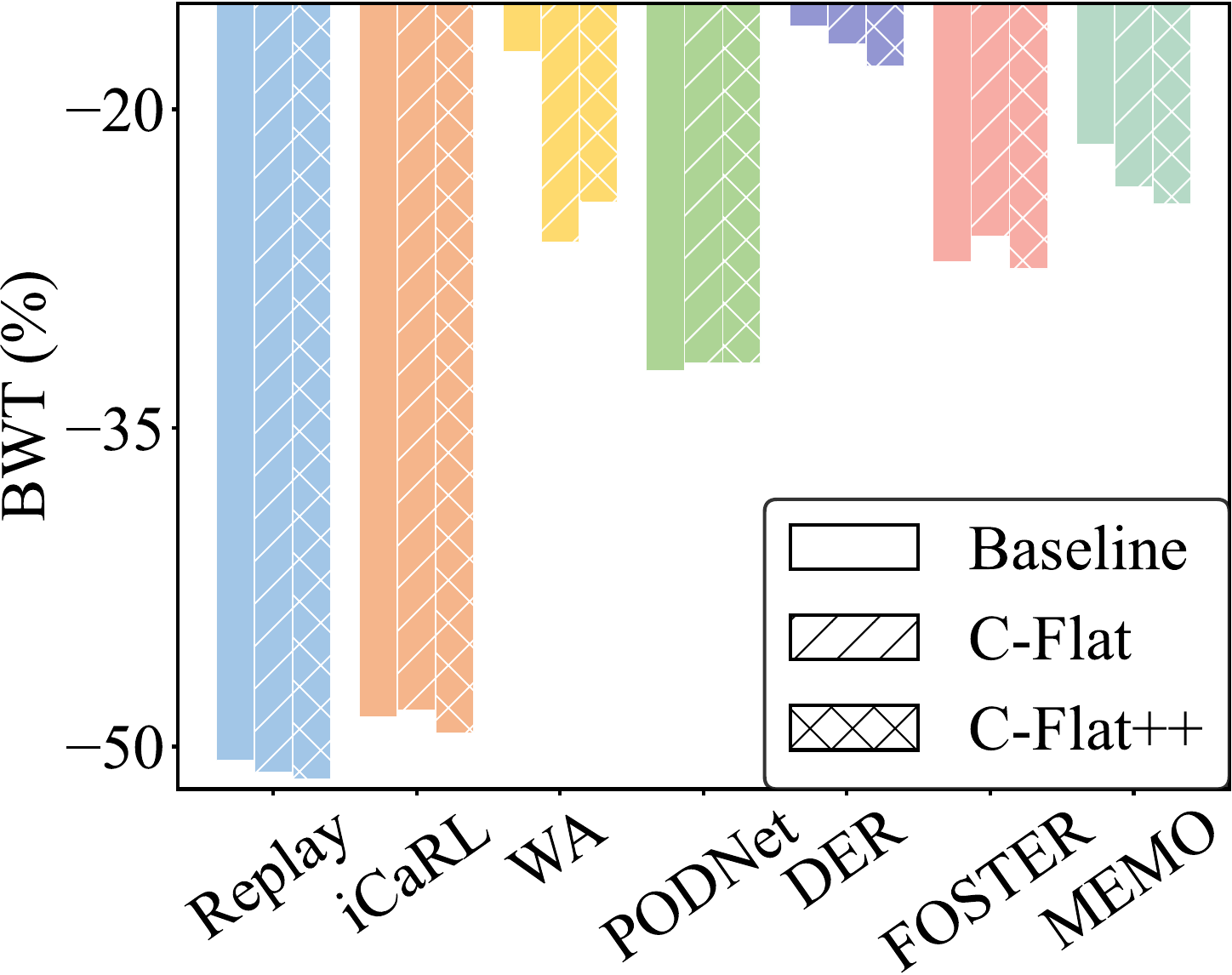}%
        \label{subfig:fwt}}
\caption{Analysis of knowledge transfer.}
\label{fig:transfer}
\vspace{-4mm}
\end{figure}

\begin{table}
\centering
\scriptsize
\caption{Analysis of old and new task accuracy when adapting last new task. RR refers to Relative Return compared to SGD.}
\label{tbl:old}
\begin{tblr}{
  row{even} = {c},
  cell{1}{1,6} = {r=2}{},
  cell{1}{3} = {c=3}{c},
  cell{3,5,7}{1} = {r=2}{},
  cell{3,5,7}{2-6} = {c},
  hline{1,3,5,7,9} = {-}{},
}
Method &        & CIFAR-100 / B0\_Inc10     &     &  & RR (C-Flat / C-Flat++) \\
       &        & SGD    & C-Flat & C-Flat++ &    \\
Replay~\cite{NEURIPS2019_fa7cdfad} & old    & 35.96  & 36.87  & 36.67    & \textcolor{red}{+2.53\%}/\textcolor{red}{+1.97\%}   \\
       & new    & 83.50  & 85.50  & 85.10    & \textcolor{red}{+2.40\%}/\textcolor{red}{+1.92\%}   \\
WA~\cite{zhao2020maintaining}    & old    & 51.49  & 51.34  & 51.83    & \textcolor{forestgreen}{-0.29\%}/\textcolor{red}{+0.66\%}  \\
       & new    & 56.60  & 69.40  & 67.60    & \textcolor{red}{+22.61\%}/\textcolor{red}{+19.43\%}   \\
MEMO~\cite{zhou2022model}   & old    & 56.63  & 59.36  & 58.56    & \textcolor{red}{+4.82\%}/\textcolor{red}{+3.41\%}   \\
       & new    & 65.40  & 68.80  & 70.00    & \textcolor{red}{+5.20\%}/\textcolor{red}{+7.03\%}
\end{tblr}
\end{table}

% Replay~\cite{NEURIPS2019_fa7cdfad}
% iCaRL~\cite{rebuffi2017icarl}
% WA~\cite{zhao2020maintaining}
% PODNet~\cite{DBLP:conf/eccv/DouillardCORV20}
% DER~\cite{DBLP:conf/cvpr/YanX021}
% FOSTER~\cite{wang2022foster}
% MEMO~\cite{zhou2022model}

As is known to all, forward and, in particular, backward transfer are desirable properties in continual learning (CL)~\cite{hadsell2020embracing, wang2024comprehensive}. Here, we thoroughly examine the performance of the proposed methods in both aspects. Forward Transfer (FWT) measures the average influence of all previously learned tasks on the current task. Backward Transfer (BT), on the other hand, quantifies how learning new tasks affects performance on earlier ones when revisited.

As shown in Figure~\ref{fig:transfer}, both C-Flat and C-Flat++ outperform baseline methods in FWT, indicating that knowledge learned by our methods facilitates the adaptation to future tasks. However, we observe that in four out of seven scenarios, C-Flat appears to suffer from more forgetting than the baselines. This is mainly because C-Flat++ establishes \textit{higher standards} of performance on previous tasks, making the drop more noticeable.
For example, on Task~1, SGD and C-Flat++ achieve 80\% and 90\% accuracy, respectively. After learning Task~2, their performance drops to 70\% and 75\%, respectively. Although C-Flat++ still performs better than SGD, the \textit{drop} in accuracy (15\% vs. 10\%) appears larger, giving the impression of heavier forgetting. 
To ensure a fair comparison, we report the average accuracy on all previously seen tasks after learning the final task, as shown in Table~\ref{tbl:old}. As observed, both C-Flat and C-Flat++ outperform the baseline methods in nearly all cases.

\subsection{Further exploration}

\begin{table}
\centering
\caption{Analysis of C-Flat++ in DIL. Bold indicates the best results and underline denotes the second-best.}
\label{tbl:domain}
\scriptsize
\begin{tblr}{
  colspec={l cccccc},
  hline{1,3,6} = {-}{},
  cell{1}{2} = {c=6}{},
  cell{1}{1} = {r=2}{},
}
Method                          & DomainNet &           &          &           &       &        \\
                                & clipart   & infograph & painting & quickdraw & real  & sketch \\
DUCT~\cite{zhou2024dual}        & 74.21     & 60.52     & 67.73    & \underline{63.50}     & 69.51 & 68.63  \\
w/ C-Flat~\cite{bian2025make}                       & \textbf{74.93}     & \textbf{60.88}     & \textbf{68.10}    & 63.30     & \underline{70.22} & \underline{69.41}  \\
w/ C-Flat++                     & \underline{74.83}     & \underline{60.71}     & \underline{67.99}    & \textbf{64.98}     & \textbf{71.13} & \textbf{70.25}  
\end{tblr}
\end{table}

While this paper primarily focuses on the effectiveness in CIL settings, the flatness-driven concepts of C-Flat naturally extend to other scenarios, such as domain-incremental learning (DIL). To demonstrate this, we evaluate our approach on DomainNet~\cite{peng2019moment}, a benchmark with significant domain shifts, following the default setup in the DUCT~\cite{zhou2024dual} repository. We set the perturbation radius $\rho$ to 0.05 here. As shown in Table~\ref{tbl:domain}, both C-Flat and C-Flat++ consistently outperform DUCT across all stages.

\section{Conclusion}

This paper presents a versatile optimization framework, C-Flat, designed to mitigate forgetting in continual learning. Empirical results demonstrate that C-Flat consistently outperforms various CL methods, highlighting its plug-and-play nature. Furthermore, analyses of Hessian eigenvalues and traces reaffirm the effectiveness of C-Flat in encouraging flatter minima that benefit CL performance. In essence, C-Flat serves as a simple yet powerful addition to the continual learning toolkit. Beyond C-Flat, we propose C-Flat++, which employs a selective, flatness-driven update mechanism based on a sharpness proxy, significantly reducing the computational overhead of C-Flat.

\section*{Acknowledgments}

This work was supported in part by the Chunhui Cooperative Research Project from the Ministry of Education of China under Grand HZKY20220560, in part by the National Natural Science Foundation of China under Grant W2433165, and in part by the National Natural Science Foundation of Sichuan Province under Grant 2023YFWZ0009.

\bibliographystyle{IEEEtran}
\bibliography{egbib.bib}

\end{document}